# Learning team-based navigation: a review of deep reinforcement learning techniques for multi-agent pathfinding

**Jaehoon Chung[1] · Jamil Fayyad[2] · Younes Al Younes[1] · Homayoun Najjaran[1]**

## Abstract

Multi-agent pathfinding (MAPF) is a critical field in many large-scale robotic applications, often being the fundamental step in multi-agent systems. The increasing complexity of MAPF in complex and crowded environments, however, critically diminishes the effectiveness of existing solutions. In contrast to other studies that have either presented a general overview of the recent advancements in MAPF or extensively reviewed Deep Reinforcement Learning (DRL) within multi-agent system settings independently, our work presented in this review paper focuses on highlighting the integration of DRL-based approaches in MAPF. Moreover, we aim to bridge the current gap in evaluating MAPF solutions by addressing the lack of unified evaluation indicators and providing comprehensive clarification on these indicators. Finally, our paper discusses the potential of model-based DRL as a promising future direction and provides its required foundational understanding to address current challenges in MAPF. Our objective is to assist readers in gaining insight into the current research direction, providing unified indicators for comparing different MAPF algorithms and expanding their knowledge of model-based DRL to address the existing challenges in MAPF.

✉ Homayoun Najjaran
najjaran@uvic.ca

Jaehoon Chung
broomshot@uvic.ca

Jamil Fayyad
jfayyad@mail.ubc.ca

Younes Al Younes
yalyounes@gmail.com

[1] Mechanical Engineering, University of Victoria, 3800 Finnerty Road, Victoria, BC V8P 5C2, Canada

[2] Mechanical Engineering, The University of British Columbia, 3333 University Way, Kelowna, BC V1V 1V7, Canada

# 1 Introduction

The Multi-Agent Pathfinding (MAPF) problem involves planning and following paths for a set of agents from their respective start positions to goal positions, without any collisions (Stern et al. 2019). Originally, the term "path planning" was typically used to find the shortest obstacle-free path for a single robot to reach its goal point. However, in multi-agent systems, there exists a trade-off between finding optimized paths and quickly generating the paths for all agents, which is crucial for their real-time execution. As a result, the research community has focused on developing algorithms for MAPF defined by Stern et al. (2019). In MAPF, finding computationally efficient and collision-free paths is a more critical consideration than finding optimized paths.

Over the last few years, there has been a considerable increase in demand for artificial intelligence that requires collaborative robotic systems in various industries (Jennings et al. 1997; Balch and Arkin 1998; Halperin et al. 1998; Fox et al. 2000; Poduri and Sukhatme 2004; Griffith and Akella 2005; Rodriguez and Amato 2010; Šišlák et al. 2010; Barer et al. 2014; Salzman and Stern 2020; Enayati et al. 2022; Khadivi et al. 2022; Ogunfowora and Najjaran 2023). In particular, leading companies have recently paid close attention to MAPF applications as MAPF often becomes a fundamental task in designing physical multi-agent systems. As the demand for cooperative and competitive environments continues to increase in many fields, the efficacy of MAPF is expected to grow even further. In many cases, the MAPF problem turns out to be the first step to be solved for the multi-robot framework, such as warehouse automation (Zhang et al. 2023), traffic management for autonomous vehicles (Dresner and Stone 2008), and coordination of multi-robot exploration (Ma 2022).

In MAPF problems, it is necessary to consider the conditions and constraints, such as assumptions about the other agents and obstacles or the objective of the problem. For example, depending on the conditions, agents can fully observe other agents and the whole environment. Although this would allow agents to generate safe and optimized paths, the computational cost will exponentially increase as the number of agents increases and it normally is inapplicable for real-world robots to have fully-observable conditions with complex settings. On the other hand, in partially-observable environments, each agent will be likely to suffer from uncertainty, which would put them at risk of collision or deadlock scenarios. Nevertheless, the research community has been trying to adapt this condition in their MAPF research and solutions due to the feasibility of real robot applications in complex environments.

Furthermore, the research community has been focusing more on decentralized execution rather than centralized one to enhance computational efficiency and expand the work to a larger scale environment. Although there has been a research based on fast planning method under centralized algorithm applicable to large scale environment (Okumura et al. 2022), the authors have recognized that it can be decentralized to address partially observability, making it a feasible solution for real-world applications. Depicted in Fig. 1, decentralized agents normally plan their paths based on two aspects: proactive planning, and reactive planning. The proactive agent's path is affected by their prediction of the next observations. This method can only provide high-quality solutions when the prediction estimation is reliable, otherwise, it can end up in risky situations. On the other hand, reactive agents correct their paths from individual observation. It can be comparatively safer and computationally more efficient but might result in suboptimal paths or failure to navigate effectively in complex scenarios.

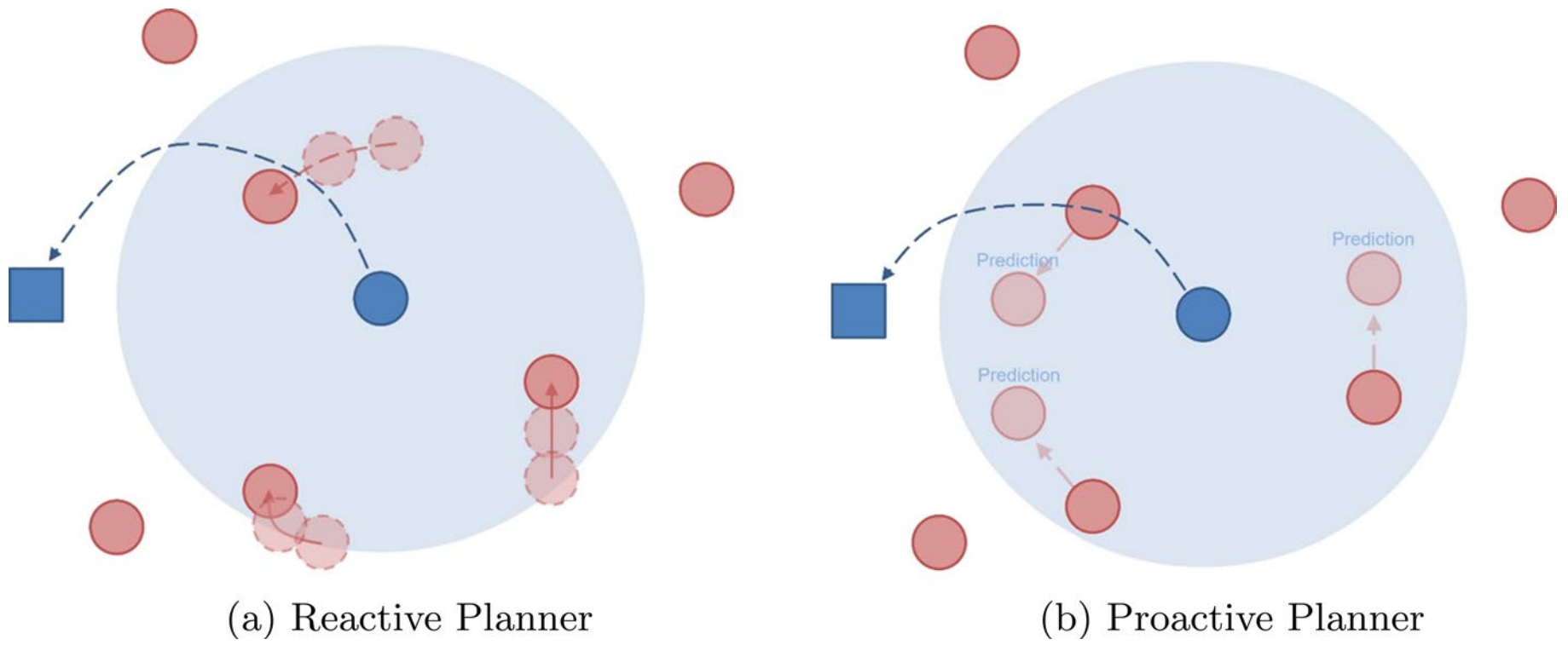

**Fig. 1** Illustration of two aspects of decentralized path planners. The blue circle represents an agent attempting to plan its path toward a goal position. The blurred blue area represents the range of the agent's observable region. The blue rectangle represents the agent's goal position. The red circles represent other agents or dynamic obstacles. **a** Reactive planning method. The agent makes decisions based on its current or historical observations of surrounding obstacles or other agents. **b** Proactive planning method. The agent predicts future observations, which affects the agent generating its path

Recent algorithms for decentralized execution which have been showing progress can mainly be divided into three approaches: bio-inspired meta-heuristics (Wong and Ming 2019; Roni et al. 2022), search-based (Li et al. 2019, 2021, 2021b; Zhang et al. 2022), and learning-based solutions (Kaduri et al. 2020; Huang et al. 2021b, a). Among the latter, deep reinforcement learning (DRL) is one of the most active areas being studied for developing safe and robust MAPF solutions. DRL-based approaches have shown promising advances in various areas due to their powerful decentralized execution, high adaptability to new contexts and uncertainty, and suitability for partially observable environments (Laurent et al. 2021; Ivanov 2022; Shojaeinasab et al. 2022; Honari and Khodaygan 2023).

In this review paper, our main purpose is to provide a comprehensive view of how DRL-based approaches have tackled the challenges in solving the large-scale complicated environments of MAPF and suggest a new research frontier for DRL-based approaches. There have been previous review works that survey overall categories of approaches in MAPF (Lin et al. 2022), how MAPF can be extended or applied to real-world (Ma 2022), and current approaches for MAPF (Stern 2019). According to these works, the research community has provided meaningful results so far but still lacks some unified evaluation indicators to compare each work even from the same benchmarks. This review suggests some meaningful evaluation indicators that we believe will provide a comparable evaluation of different algorithms. In addition, DRL algorithms have been facing some challenges in a long-range planning problem (Eysenbach et al. 2019), and they are not guaranteed to provide generalized solutions in highly non-stationary environments (Zhang et al. 2021). Still, we believe DRL-based approaches are going to be one of the breakthroughs for MAPF so it would be worth it to focus on how DRL-based approaches have been developing. The summary of the major contributions of this paper is as follows.

- Clarify meaningful evaluation indicators that can be unified to evaluate the effectiveness of algorithms in large-scale, complex MAPF environments.
- Provide a comprehensive review and classification of the recent DRL-based approaches for MAPF.

- Introduce the concept of reactive planner and proactive planner to map it with model-free and model-based DRL planners for MAPF. This concept gives some inspiration why we believe model-based methods might lead to advancement in MAPF.
- Bridge the recent DRL-based approaches as an intermediate stage between model-free methods and model-based methods. Some backgrounds and prerequisites for model-based DRL would be discussed in Sect. 4.

The rest of this paper is organized as follows: Sect. 2 provides an overview of MAPF research to help readers better understand the main objective of MAPF and bridge DRL algorithms with the problem. Section 3 provides the recent advancement of DRL-based approaches for MAPF. In Sect. 4, we provide a foundational overview of key aspects in model-based reinforcement learning and discuss the promise of it for MAPF solutions. Finally, we give brief conclusions in Sect. 5.

## 2 Background of MAPF approaches

In this section, we present an overview of MAPF and discuss its problem formulation. Additionally, we provide several recent approaches to MAPF algorithms. Moreover, we highlight the advantages of incorporating DRL approaches into MAPF and present the mathematical foundation of DRL-based algorithms. Finally, we discuss several evaluation indicators and conditions for evaluating the performance and conditions of MAPF algorithms.

### 2.1 Overview of MAPF

The MAPF problem consists of a set of agents, obstacles, and target points in a given map. The main objective is to compute collision-free paths where all agents can reach assigned targets from their start locations. Once multiple feasible paths are found, the problem can then be regarded as an optimization problem since the aim is to identify the paths that optimize a given objective function. From a classical point of view (Stern 2019), MAPF with $N$ agents, $M$ dynamic, and $K$ static obstacles can be defined as a tuple: $\langle \boldsymbol{\mathcal{A}}, \boldsymbol{\mathcal{G}}, \boldsymbol{\mathcal{O}}_{dyn}, \mathscr{O}_{static} \rangle$, where $\boldsymbol{\mathcal{A}} = (\mathcal{A}_1, \ldots, \mathcal{A}_N)$ is a set of $N$ agents, $\boldsymbol{\mathcal{G}} = (\boldsymbol{\mathcal{V}}, \boldsymbol{\mathcal{E}})$ is a graph that represents the whole map environment, $\boldsymbol{\mathcal{O}}_{dyn} = (\mathcal{O}_1, \ldots, \mathcal{O}_M)$ is a set of $M$ dynamic obstacles, and $\mathscr{O}_{static} = (\mathscr{O}_1, \ldots, \mathscr{O}_K)$ is a set of static obstacles. Every agent $\forall \mathcal{A}_n \in \boldsymbol{\mathcal{A}}$ is characterized by a tuple $\langle \mathfrak{s}_n, \mathfrak{g}_n \rangle$ where $\mathfrak{s}_n$ and $\mathfrak{g}_n$ are agent $\mathcal{A}_n$'s start location and goal location, respectively. The vertices $\forall \mathcal{V}_i \in \boldsymbol{\mathcal{V}}$ can be occupied by any agents $\forall \mathcal{A}_n \in \boldsymbol{\mathcal{A}}$ or any obstacles $\forall \mathcal{O}_m \in \boldsymbol{\mathcal{O}}_{dyn}$ and $\forall \mathscr{O}_k \in \mathscr{O}_{static}$. The agents and dynamic obstacles can move from $\mathcal{V}_i$ to $\mathcal{V}_j, \forall \mathcal{V}_i, \mathcal{V}_j \in \boldsymbol{V}$ through $\mathcal{E}_{ij} \in \boldsymbol{\mathcal{E}}$. Time is assumed to be discretized into time steps so each agent can decide on a single action where to move through an adjacent edge for every time step. When there exists a joint sequence of actions $\boldsymbol{\Pi} = (\Pi_1, \ldots, \Pi_N)$ where $\Pi_n = (a_0^{(n)}, \ldots, a_f^{(n)}) \in \boldsymbol{\Pi}$ satisfies $a_f^{(n)}(\cdots a_1^{(n)}(a_0^{(n)}(\mathfrak{s}_n))) = \mathfrak{g}_n$ without any deadlock scenario and collision, we say this is the valid solution of MAPF. The conditions for a valid solution of MAPF should satisfy:

- No deadlock scenario: $\exists T \in \mathbb{Z}^{+}$ s.t. $f = T$.
- No collision on vertices: $\forall \Pi_n, \Pi_{n'} \in \mathbf{\Pi}, \forall 0 \le t \le f, a_t^{(n)}(\cdots a_0^{(n)}(\mathfrak{s}_n)) \neq a_t^{(n')}(\cdots a_0^{(n')}(\mathfrak{s}_{n'})), a_t^{(n)}(\cdots a_0^{(n)}(\mathfrak{s}_n)) \notin (\mathcal{O}_{dyn} \cup \mathcal{O}_{static})$.
- No collision on edges: $\forall \Pi_n, \Pi_{n'} \in \mathbf{\Pi}, \forall 0 \le t \le f, \forall \mathcal{O}_m \in \mathcal{O}_{dyn}, (a_t^{(n)}(\cdots a_0^{(n)}(\mathfrak{s}_n)), a_t^{(n')}(\cdots a_0^{(n')}(\mathfrak{s}_{n'}))) \neq (a_{t+1}^{(n)}(\cdots a_0^{(n)}(\mathfrak{s}_{n'})), a_{t+1}^{(n')}(\cdots a_0^{(n')}(\mathfrak{s}_n))), (a_t^{(n)}(\cdots a_0^{(n)}(\mathfrak{s}_n)), \mathcal{O}_m(t)) \neq (a_{t+1}^{(n)}(\cdots a_0^{(n)}(\mathfrak{s}_n)), \mathcal{O}_m(t+1))$, where $(a_t^{(n)}(\cdots a_0^{(n)}(\mathfrak{s}_n)), a_t^{(n')}(\cdots a_0^{(n')}(\mathfrak{s}_{n'})))$, $(a_t^{(n)}(\cdots a_0^{(n)}(\mathfrak{s}_n)), \mathcal{O}_m(t)) \in \mathcal{E}$ and $\mathcal{O}_m(t)$ is the vertex occupied by $\mathcal{O}_m(t)$ at time step $t$.

When more than one valid solution exists, MAPF solution can be optimized by given objective functions. The two most widely used objective functions for MAPF are *Makespan* and *Sum of Costs*. The *Makespan*, $M(\mathbf{\Pi})$, is the number of total time steps required for all agents to reach their target locations:

$$M(\mathbf{\Pi}) = \max_{1 \le k \le N} |\Pi_k| \tag{1}$$

The *sum of costs*, $SOC(\mathbf{\Pi})$, is the total required number of actions for all agents to reach their target locations:

$$SOC(\mathbf{\Pi}) = \sum_{k=1}^{N} |\Pi_k| \tag{2}$$

Figure 2 depicts the example of conditions required to be considered for deriving a valid solution and of objective functions.

As discussed in Sect. 1, researchers are currently considering decentralized approaches to expand the MAPF problem to a larger scale. In the following Sect. 2.1.1, we present a classification for some of the recent decentralized MAPF approaches.

### 2.1.1 MAPF Approaches

Over the past few decades, plenty of approaches have been proposed for solving MAPF. Most of these approaches have been reviewed and classified in a plethora of review papers (Foead et al. 2021; Bertolini 2022; Lin et al. 2022; Tjiharjadi et al. 2022). The recent research direction, however, is mainly focused on decentralized executions that allow deploying algorithms in large-scale environments. Hence, our focus here is to provide a comprehensive survey that covers recent promising decentralized approaches in the field.

Figure 3 presents the classification of decentralized MAPF approaches, which are divided into the following categories: Learning-based, Bio-inspired Meta-Heuristics, and Search-based approaches.

Search-based approaches refer to approaches that utilize search algorithms to find optimal solutions for MAPF. The early proposed search-based algorithms stem from a variant of A* (Standley 2010; Wagner and Choset 2015; Ravankar et al. 2017; Li et al. 2022; Zagradjanin et al. 2019; Serpen and Dou 2015), which finds the shortest path between the start and goal nodes by considering the cost of reaching the current node and an estimate of the remaining cost to reach the goal node. However, A* often suffers from large-scale MAPF because not only the state space but also the branching factor of each state exponentially increases as the number of agents grows.

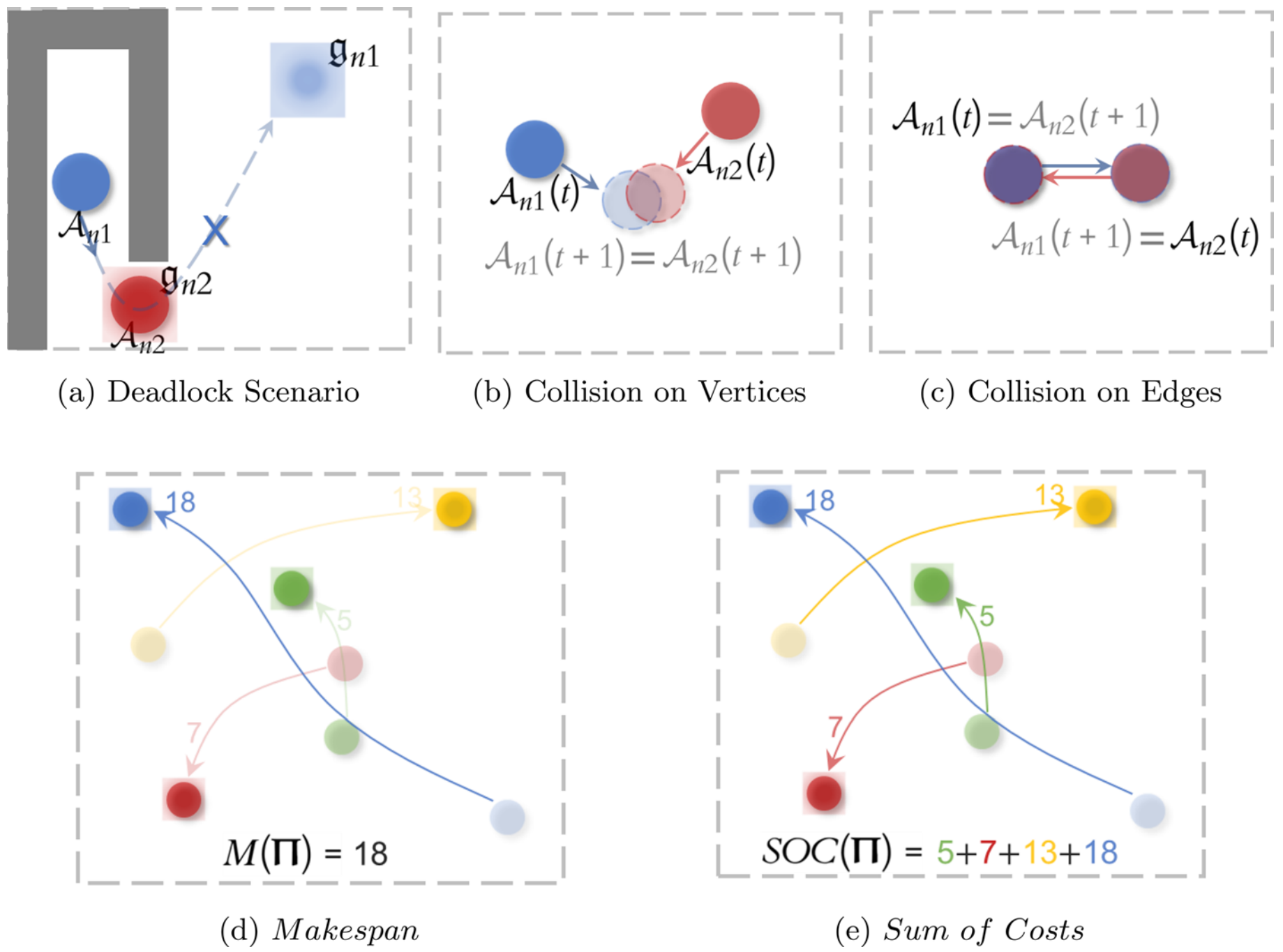

**Fig. 2** Illustration of conditions resulting in failure for MAPF (**a**–**c**), and two widely used objective functions for optimizing the solution (**d**–**e**). The circles represent the agents and the corresponding color of rectangles are the respective goal positions. Each agent generates its own path toward the goal position. The numbers in **d** and **e** are the number of time steps required for the corresponding color of agents to reach the goal position

Recent search-based research in decentralized MAPF which has drawn a lot of attention is adapting the hierarchical mechanism that consists of two-level MAPF resolution. One of the promising methods that have been widely adopted as the root approach is the Conflict-based search (CBS)-based approach (Sharon et al. 2015). CBS uses a two-level search, the higher-level search manages collisions between agents, and the lower-level search finds paths for multi-agents in the context of the high-level constraints. It was first developed as a centralized optimal MAPF solver, but its flexibility to combine with other algorithms allows it to be applied to decentralized MAPF (Okumura et al. 2023; Kottinger et al. 2022).

Bio-inspired meta-heuristic algorithms also have emerged as one of the leading solutions for MAPF. Inspired by the biological processes in nature, such as Particle Swarm Optimization (PSO), Ant Colony Optimization (ACO), and Genetic Algorithm (GA), They are used to solve optimization problems that are difficult or impossible to find optimal or sub-optimal solutions. Although it may not guarantee a globally optimal solution, it can provide a sufficiently good solution with incomplete or imperfect information with efficient computational cost (Bianchi et al. 2009).

Inspired by the social behavior of bird flocks or fish schools, PSO solves an optimization problem by having a population of candidate solutions that move around in the search space to find optimal solutions. Recently it's been being adapted as a planner for decentralized MAPF (Das et al. 2016; Wang et al. 2019; Ahmed et al. 2021; Tang et al. 2020). ACO, on the other hand, is a probabilistic technique inspired by the foraging behavior of ant colonies secreting pheromone-based communication to find the shortest path to the food source. ACO can also be used to develop decentralized MAPF (Seyyedabbasi and Kiani 2020). GA is based on the evolution mechanism, the process that drives biological evolution. In GAs, genetic operators which follow the process of natural selection and genetics select candidate solutions to search for the optimal solution. due to their ability to find global optima and high parallelism, They have been being used for decentralized MAPF solutions by introducing co-evolution mechanism (Qu et al. 2013; Nazarahari et al. 2019; Trudeau and Clark 2019).

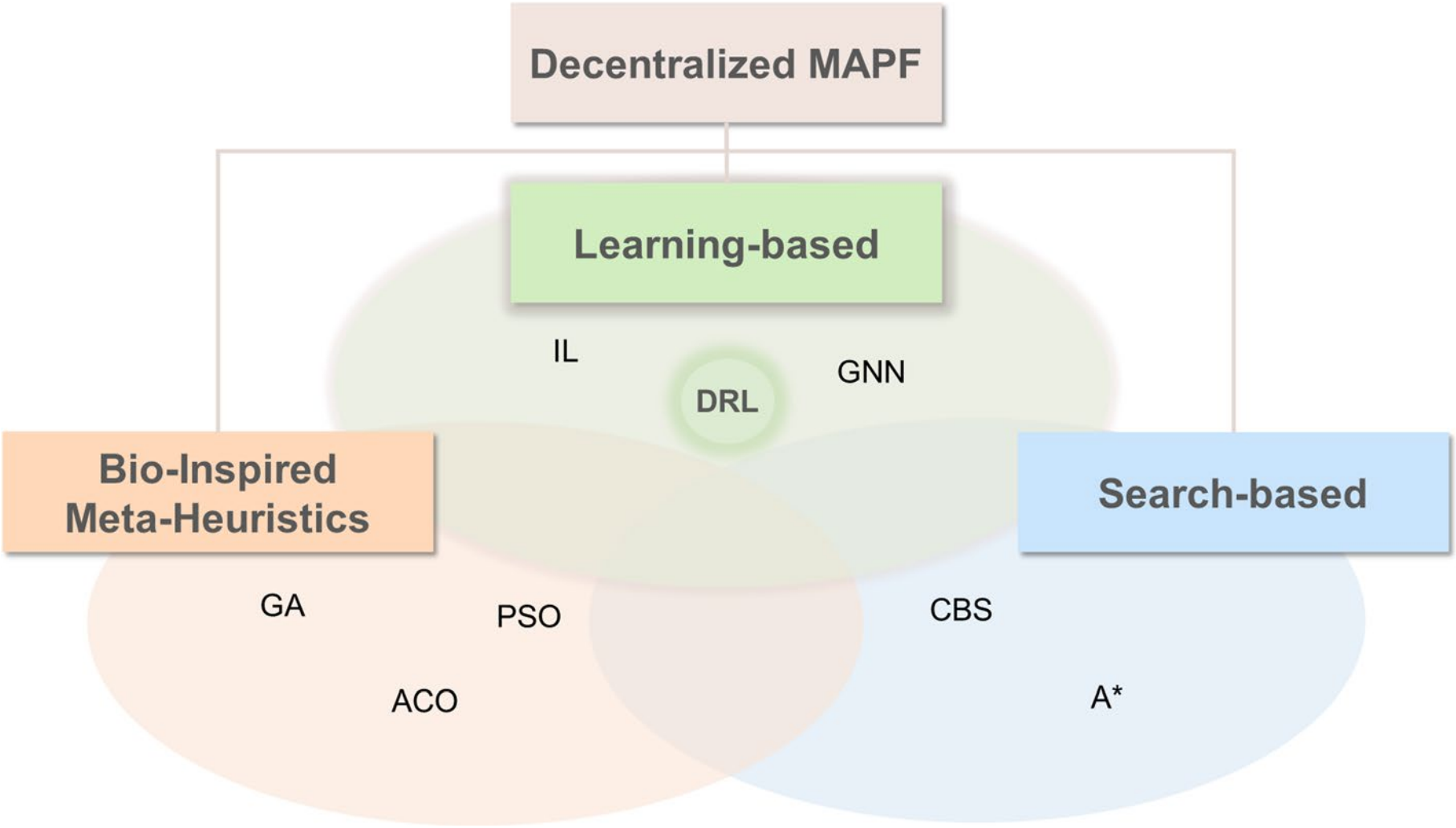

**Fig. 3** Three main categories in decentralized approaches for MAPF

Although Bio-inspired Meta-Heuristics approaches excel at exploring the search space for optimal solutions, they might suffer from computational cost as the dimensions of the problem increase significantly (Hussain et al. 2019). To expand MAPF to a large environment with a number of robots, it is inevitable to confront the challenge of dimensionality. In that sense, Learning-based approaches can be effective to address such challenges for MAPF.

In particular, the recent advances in deep learning and reinforcement learning have recorded sublime success in decentralized MAPF. Combined together, DRL has gotten rapid traction and is currently thriving with accomplishments mastering the complexity of MAPF by employing neural networks as function approximators (Arulkumaran et al. 2017; Gronauer and Diepold 2022).

## 2.2 DRL-based approaches in MAPF

This section provides insights into DRL to help new researchers in this field in building a solid foundation. In the subsequent sections, we will delve into the strengths of DRL-based approaches in MAPF, the DRL-based MAPF framework, a variety of DRL algorithms, and multi-agent DRL algorithms.

#### 2.2.1 Effectiveness of using DRL-based approaches in MAPF

The recent advancements in centralized systems have enabled more organized and improved control of multi-agent coordination. However, the inherent complexity and scale of these systems make it hard to handle vast and intricate environments due to computational challenges. To overcome this issue, researchers have proposed various solutions, notably decentralized approaches, which have garnered considerable attention and are actively investigated. Among these approaches, DRL has emerged as a promising technique in decentralized systems. DRL exhibits adapting capabilities in interactive environments and varying conditions through iterative learning via a trial-and-error approach. Employing neural networks as function approximators allow individual agents to acquire optimal decision-making policies based on their local observations, bypassing the need for frequent computation for communication with other agents or dynamic obstacles.

DRL-based approaches can adapt without requiring explicit knowledge of the environment's dynamics. They can facilitate efficient coordination and cooperation among agents, eliminating the need for explicit coordination mechanisms (Li et al. 2022b; Chen et al. 2022a). Furthermore, the combination of DRL-based approaches with other techniques has yielded powerful solutions that leverage the respective strengths of each method. For example, Sehgal et al. (2019) enhanced the learning process of DRL by integrating GA to tune the DRL parameters, resulting in more effective learning and improved performance. Additionally, the integration of an optimal search algorithm with DRL encourages agents to optimize the disparity between optimal solutions while managing uncertainties (Wang et al. 2020). Consequently, DRL-based approaches offer a promising avenue for addressing the challenges associated with executing real-time implementations in complicated and huge environments.

#### 2.2.2 DRL-based MAPF framework

MAPF can basically be formulated as a Markov Game (MG), an extension of the Markov Decision Process (MDP) framework to capture decision-making in multi-agent settings (Littman 1994). An MDP is represented by a 5-tuple $< \mathbf{S}, \mathbf{A}, \mathscr{T}, \mathfrak{R}, \gamma >$, where

- $\mathbf{S}$ is the state space,
- $\mathbf{A}$ is the action space,
- $\mathscr{T} : \mathbf{S} \times \mathbf{A} \rightarrow P(\mathbf{S})$ is the state transition distribution function,
- $\mathfrak{R} : \mathbf{S} \times \mathbf{A} \rightarrow \mathbb{R}$ is the reward function, and
- $\gamma \in [0, 1]$ is the discount factor.

The objective of an MDP is to select actions that maximize the expected long-term cumulative reward, considering an unknown transition function $\mathscr{T}$ and the reward function $\mathfrak{R}$. The agent learns a behavior policy $\pi : \mathbf{S} \rightarrow P(\mathbf{A})$ that optimizes the expected performance $G$ throughout the learning process. The performance is defined as the expected cumulative reward over the initial state distribution $\rho_0$:

$$G = E_{s_0 \sim \rho_0, s_t \sim \mathscr{T}, a_t \sim \pi}\Big[\sum_{t=0}^{\infty} \gamma^t \mathfrak{R}(s_t, a_t)\Big]. \tag{3}$$

To optimize $G$ from a given state $s$, we can utilize the state-value function $V_\pi : \mathbf{S} \rightarrow \mathbb{R}$ which estimates the desirability of being in a particular state $s$ under policy $\pi$:

$$V_\pi(s) = E_{s_t \sim \mathscr{T}, a_t \sim \pi}\Big[\sum_{t=0}^{\infty} \gamma^t \mathfrak{R}(s_t, a_t) \mid s_0 = s\Big]. \tag{4}$$

By updating the policy $\pi$ to maximize the state-value function, the agent iteratively optimizes the expected performance $G$. Similarly, we can employ the action-value function $Q_\pi : \mathbf{S} \times \mathbf{A} \rightarrow \mathbb{R}$, also known as Q-function, which estimates the value of selecting a specific action $a$ in a given state $s$ under policy $\pi$:

$$Q_\pi(s, a) = E_{s_t \sim \mathscr{T}, a_{t>0} \sim \pi}\Big[\sum_{t=0}^{\infty} \gamma^t \mathfrak{R}(s_t, a_t) \mid s_0 = s, a_0 = a\Big]. \tag{5}$$

In DRL, the policy $\pi$ or value functions are represented by neural networks.

In an $N$ multi-agent MG, in comparison to an MDP, we need to consider the involvement of multiple agents $\boldsymbol{\mathcal{A}}$, denoted as we discussed in Sect. 2.1. Moreover, the action space $\mathbf{A}$ expands to a joint action space $\mathbf{A}_{1...N} = \mathbf{A}_1 \times \cdots \times \mathbf{A}_N$ and the reward function $\mathfrak{R}$ to a set of reward functions $\boldsymbol{\mathfrak{R}} = \{\mathfrak{R}_1, ..., \mathfrak{R}_N\}$, where $\mathfrak{R}_n : \mathbf{S} \times \mathbf{A}_n \rightarrow \mathbb{R}$. This expansion necessitates the adjustment of the state transition distribution function, resulting in $\mathscr{T} : \mathbf{S} \times \mathbf{A}_{1...N} \rightarrow P(\mathbf{S})$. Eventually, MG can be represented by the tuple $< \boldsymbol{\mathcal{A}}, \mathbf{S}, \mathbf{A}_{1...N}, \mathscr{T}, \boldsymbol{\mathfrak{R}}, \gamma >$.

In most cases, agents lack direct observation of the underlying state for optimal action selection, necessitating a mapping from observation history or belief states to actions. To address this challenge, some studies have extended the MG to a Partially Observable Markov Game (POMG) (Ma et al. 2021a). POMG targets at partially observable environments and is represented as a tuple $< \boldsymbol{\mathcal{A}}, \mathbf{O}_{1...N}, \mathbf{S}, \mathbf{A}_{1...N}, \mathscr{T}, \boldsymbol{\mathfrak{R}}, \gamma >$, where

- $\mathbf{O}_{1...N} = \mathbf{O}_1 \times \cdots \times \mathbf{O}_N$ is the joint observation space where $\mathbf{O}_n$ is agent $n$'s observation space,
- $\mathscr{T} : \mathbf{S} \times \mathbf{A}_{1...N} \rightarrow P(\mathbf{S} \times \mathbf{O}_{1...N})$ is the state transition and observation distribution function.

In addition to POMG, there have been several approaches that solved MAPF with a decentralized Partially Observable Markov decision process (dec-POMDP), offering simpler modeling of the problem (Guan et al. 2022; Fan et al. 2020). Unlike POMG, which formulates MAPF in respect of the entire environment, dec-POMDP focuses on an individual agent's perspective of the environment in a decentralized manner by adding a term for observing other agents and the environment. dec-POMDP can be represented by the tuple $< \mathbf{O}, \mathbf{S}, \mathbf{A}, \mathscr{T}, \mathfrak{R}, \varphi, \gamma >$, where

- $\varphi$ denotes conditional observation probability distribution, reflecting the added term.

This formulation proves advantageous for managing large-scale MAPF problems by reducing the dimensionality of the MAPF framework. Figure 4 depicts the frameworks pertinent to MAPF.

### 2.2.3 Categorization of DRL algorithms

DRL-based approaches, in general, are a specific type of model-free methods that employ neural networks to approximate the value function or directly represent the policy. These methods depend on experiential learning rather than relying on a predefined model of the environment, which includes state transitions $\mathscr{T}$ as discussed in Sect. 2.2.2. The networks of these methods take the environmental states or observations as input, with the option to incorporate or exclude the corresponding actions. Through reinforcement learning algorithms, The networks are trained to estimate value functions or directly derive actions. Shown in Fig. 5, DRL-based approaches can be broadly classified into two categories based on the roles of the networks: value-based approximation methods, and policy-gradient methods.

Value-based approximation methods focus on learning the optimal value function which we discussed in Sect. 2.2.2 (Eqs. 4, 5). The objective of these methods is to iteratively update the network parameters for either value function $V_\pi$ or $Q_\pi$ towards the optimal value function $V_{\pi^*}$ or $Q_{\pi^*}$. One widely adopted value-based method is Deep Q-Network (DQN) (Mnih et al. 2015) that utilizes deep neural networks to approximate the Q-function $Q_\pi$, as defined in Eq. 5. During the training process, experience replay $D$ is employed to randomly sample a minibatch $E = \{e_1, ..., e_I\}$, where $e_i = (s_i, a_i, r_i, s'_i)$, $\forall i \in \{1, ..., I\}$ is a (state, action, reward, next state) pair extracted from the agent's experience. This technique reduces the correlation between minibatch samples, helping to stabilize the convergence of the training process

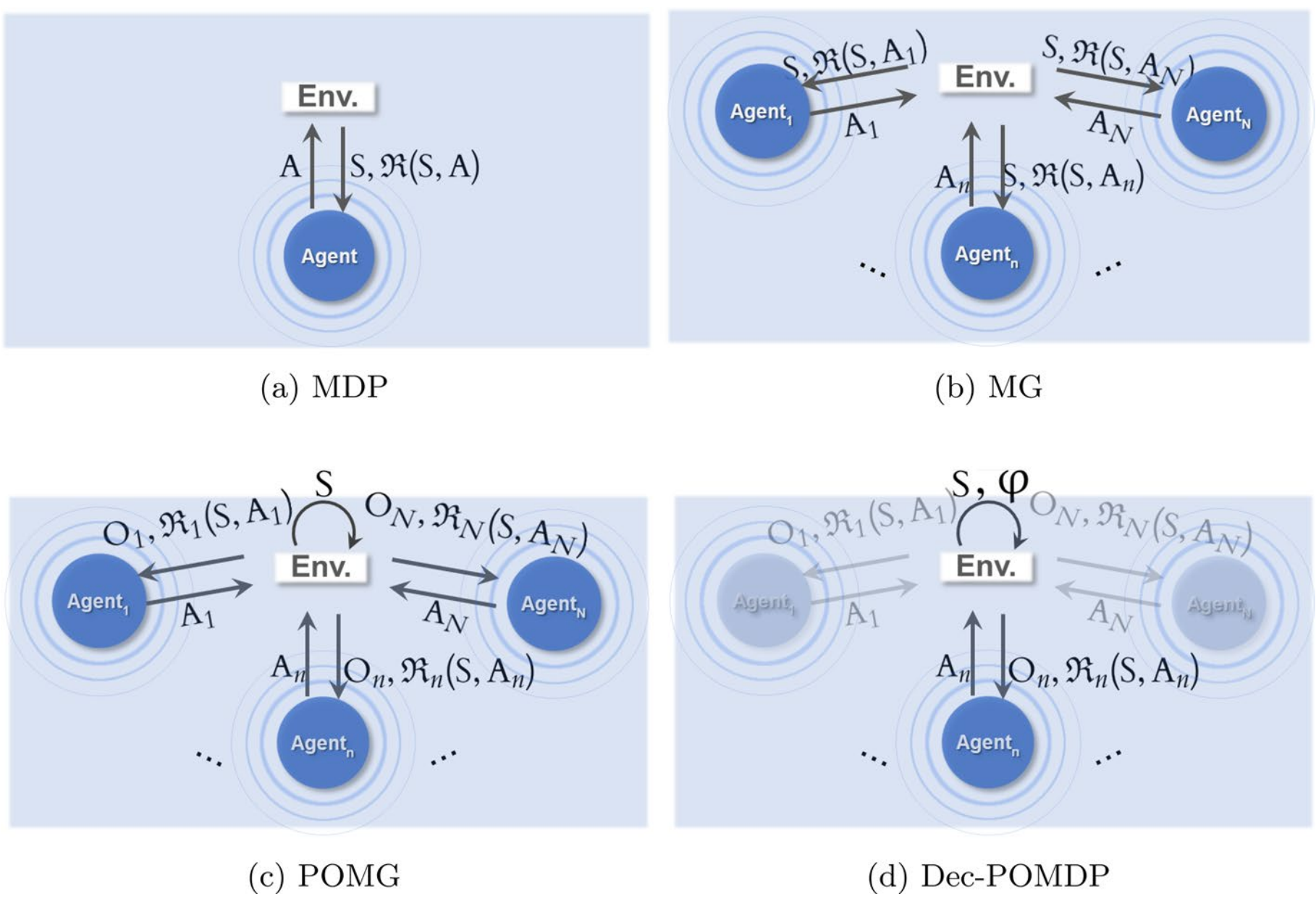

**Fig. 4** Schematic illustration of MAPF framework formulation. In **a**, the agent observes the state S and receives reward $\mathfrak{R}(S, A)$ from the environment, after outputting the action A. In **b**, multiple agents receive the environmental state S and individual rewards $(\mathfrak{R}_1(S, A_1), ..., \mathfrak{R}_N(S, A_N))$, after outputting respective actions $(A_1, ..., A_N)$ at the same time. In **c**, the agents cannot observe the state and receive individual observations $(O_1, ..., O_N)$ instead. In **d**, the individual agent's perspective of the environment is more focused with a conditional observation probability distribution ($\varphi$)

by mitigating the impact of non-stationary target distributions. The network parameters $\boldsymbol{\theta}$ of DQN are updated by minimizing the loss function, which calculates the mean-square error (MSE) between the Q-values obtained from the experience replay minibatch and the corresponding target Q-values:

$$\boldsymbol{\theta} = \arg\min_{\boldsymbol{\theta}'} \sum_{i=1}^{I} \left( r_i + \gamma \max_{a'} Q(s_i', a'; \tilde{\boldsymbol{\theta}}) - Q(s_i, a_i; \boldsymbol{\theta}') \right)^2, \tag{6}$$

where $\gamma$ is the discount factor, and $\tilde{\boldsymbol{\theta}}$ represents the target Q-network parameters, which are updated only after a fixed number of steps and are held constant between individual updates. Several variants of DQN have been proposed to enhance its performance: Double-DQN (DDQN) (Van Hasselt et al. 2016), Dueling-DQN (Wang et al. 2016), and Rainbow method (Hessel et al. 2018). In DDQN, two separate Q-value estimators, namely the selection network for the action selection and the target network for the action evaluation, are used to update each other. This could address the maximization bias and overestimation issues of DQN. On the other hand, Dueling-DQN employs two separate streams which estimate the state-value and the action advantages, while sharing the network parameters. This architecture has shown improved estimation of the state-values and more stable and effective learning ability than DQN. Besides, Hessel et al. (2018) introduced various techniques

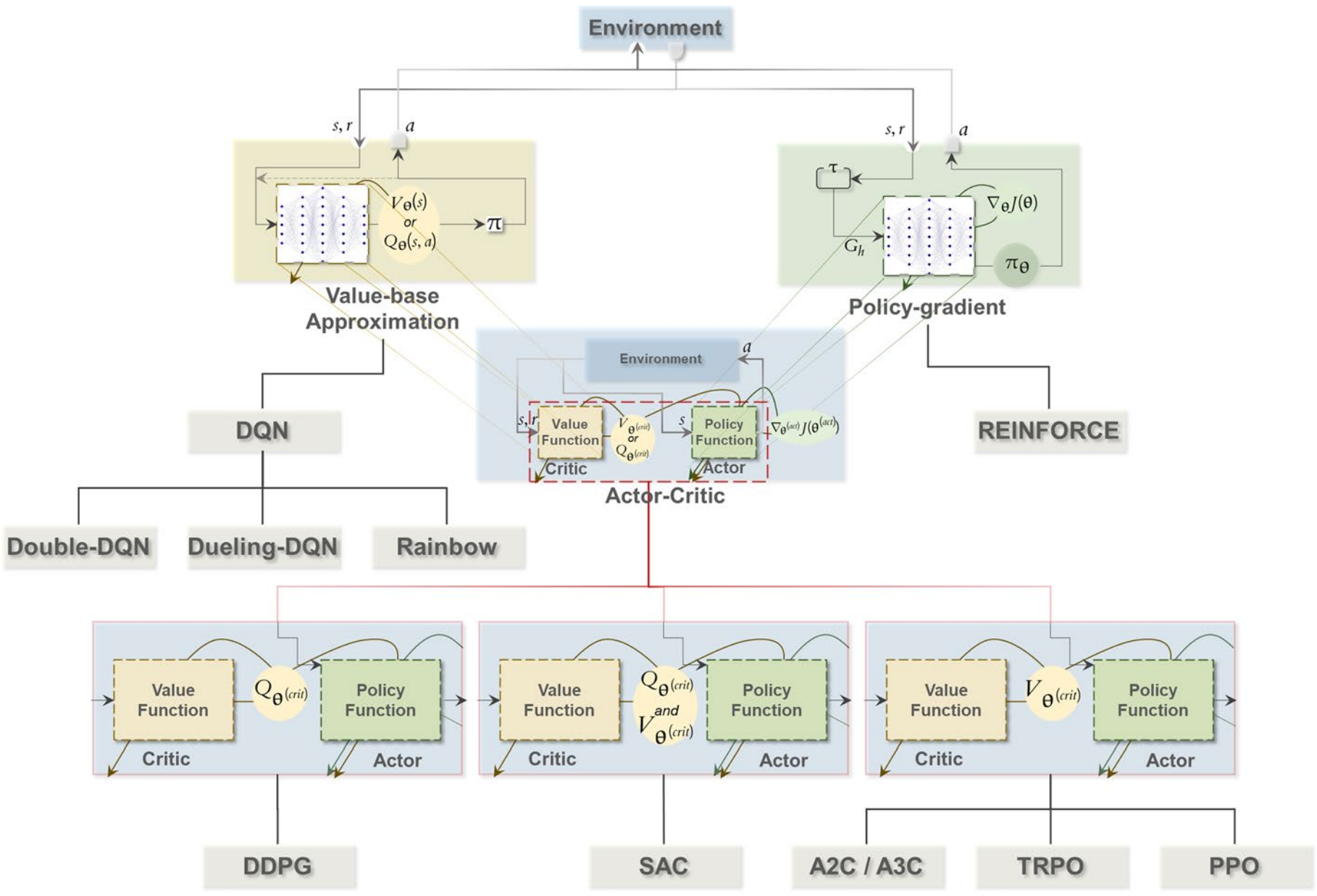

**Fig. 5** An illustrative overview of Sect. 2.2.3. Value-based approximation method uses neural networks for learning the optimal value function. DQN is one representative algorithm of the value-based approximation method and its variants are Double-DQN, Dueling-DQN, and Rainbow. Policy gradient methods uses neural networks to directly learn an optimal policy. REINFORCE algorithm is the representative algorithm of the policy gradient methods. Actor-critic methods use two neural network architectures, critic-network for learning the value function and actor-network for learning the policy. It has three types of variants. In DDPG, the critic-network learns the Q function. In A2C/A3C, TRPO, and PPO, the critic-network learns the state-value function. In SAC, the critic-network learns both the Q function and state-value function

to enhance the performance, which were all integrated into another extension, Rainbow. Rainbow has emerged as a state-of-the-art among DQN variants. While value-based methods are effective in large state spaces, they face limitations when applied to large or continuous action spaces due to the substantial computational requirements for policy updates. Moreover, the deterministic nature of the policy derived by DQN may not be suitable for situations that require a stochastic policy.

Alternatively, policy-gradient methods model the policy $\pi$ directly from the network, instead of approximating the value function, s.t.:

$$\pi_\theta(a|\ s) = P(a|s;\boldsymbol{\theta}), \tag{7}$$

for any given state $s$ and action $a$. One of the most well-known policy-gradient algorithms, the REINFORCE algorithm (Williams 1992), utilizes Monte Carlo Sampling (Hastings 1970) and computes the value function based on trajectories $\tau = (s_0, a_0, r_1, \cdots, a_H, r_{H+1}, s_{H+1})$, where $H \in \mathbb{Z}$ can take any value between 0 and the last time step of the episode. This enables the network to be updated in continuous tasks or from a portion of the entire episode. The REINFORCE algorithm updates the policy network to maximize the value function of the sampled trajectories. In contrast to value-based methods, The REINFORCE algorithm calculates the value function with the expected cumulative rewards for a trajectory and the policy network. The expected cumulative reward for trajectory $\tau$ is denoted as:

$$G_h = \sum_{t=h+1}^{H+1} \gamma^{t-h-1} r_h, \tag{8}$$

and the return for a trajectory $\tau$ is denoted as $R(\tau) = (G_0, ..., G_H)$. By performing gradient ascent on the expected return $J(\boldsymbol{\theta})$ and updating the policy network parameters, the network gradually adjusts its parameters to reach a global or local maximum of $J(\boldsymbol{\theta})$, s.t.:

$$J(\boldsymbol{\theta}) = \sum_{\tau} P(\tau;\boldsymbol{\theta})R(\tau) = E_{\pi_\theta}[R(\tau)]. \tag{9}$$

The gradient of $J(\boldsymbol{\theta})$ with respect to the policy parameters $\boldsymbol{\theta}$ then is computed as:

$$\nabla_\theta J(\boldsymbol{\theta}) = E_{\pi_\theta}\Big[R(\tau)\nabla_\theta \log \pi_\theta(\tau)\Big] = \sum_{h=0}^{H} G_h \frac{\nabla_\theta \pi_\theta(a_h|s_h)}{\pi_\theta(a_h|s_h)} \tag{10}$$

The policy network parameters $\boldsymbol{\theta}$ at time step $t+1$ are obtained by:

$$\boldsymbol{\theta}_{t+1} = \boldsymbol{\theta}_t + \alpha \nabla_\theta J(\boldsymbol{\theta}). \tag{11}$$

While this method exhibits a good convergence property, it still suffers from slow learning due to the high gradient variance of $R_\tau$. This issue can be addressed by introducing a baseline $b$ to the Eq. 10, s.t.:

$$\nabla_\theta J(\boldsymbol{\theta}) = \sum_{h=0}^{H} (G_h - b) \frac{\nabla_\theta \pi_\theta(a_h|s_h)}{\pi_\theta(a_h|s_h)}, \tag{12}$$

where $b$ should be independent of the policy parameters to keep the gradient estimate unbiased.

This idea establishes the notion of actor-critic (AC) methods by substituting the value-based network for the baseline $b$ within the framework of policy-gradient methods. AC methods leverage the strengths of value-based methods while addressing the high gradient variance issue in policy-gradient methods. They achieve this by incorporating a critic-network that approximates the value function and an actor-network for directly updating the policy. The critic-network can be utilized in two ways to reduce the gradient variance: by approximating the Q-function $Q_{\theta}$or by approximating the state-value function $V_{\theta}$.

When the critic-network approximates the Q-function, the actor-network parameters $\boldsymbol{\omega}$ are updated by considering all possible actions in a given state, as expressed by the following equation:

$$\boldsymbol{\omega}_{t+1} = \boldsymbol{\omega}_t + \alpha \sum_{\hat{a}} Q_{\theta}(s, \hat{a}) \nabla_{\boldsymbol{\omega}_t} \pi(a|s;\boldsymbol{\omega}_t). \tag{13}$$

This concept has spurred the development of advanced algorithms, such as Deep Deterministic Policy Gradient (DDPG) (Lillicrap et al. 2015). DDPG directly learns a deterministic policy with samples extracted from the experience replay to aim problems with continuous action spaces.

On the other hand, when the critic-network approximates the state-value function, it is used to derive the advantage function $A_{\theta}$, which replaces the $(G_h - b)$ term in Eq. 12 through gradient bootstrapping:

$$A_{\theta}(s, a) = r + \gamma V_{\theta}(s') - V_{\theta}(s). \tag{14}$$

The actor-network is then updated as follows:

$$\boldsymbol{\omega}_{t+1} = \boldsymbol{\omega}_t + \alpha A_{\theta}(s, \hat{a}) \frac{\nabla_{\boldsymbol{\omega}_t} \pi(a|s;\boldsymbol{\omega}_t)}{\pi(a|s;\boldsymbol{\omega}_t)}. \tag{15}$$

This approach is also known as Advantageous Actor-Critic (A2C). When multiple workers interact with separate copies of the environment in parallel and independently update their own networks, the framework becomes Asynchronous Advantage Actor-Critic (A3C) (Mnih et al. 2016). A3C enables more extensive exploration of the state-action space in significantly less time, as each agent continues independent exploration and updates the global network. Trust Region Policy Optimization (TRPO) (Schulman et al. 2015a) and Proximal Policy Optimization (PPO) (Schulman et al. 2017) are another advanced algorithms that are similar to A2C. TRPO enforces a trust region constraint on policy updates to ensure stability and reliable updates to the actor-network. On the other hand, PPO optimizes the policy by constraining the maximum change in the policy to prevent destabilization, employing a surrogate objective function.

Additionally, a recent paper by Haarnoja et al. (2018) introduces the concept of maximum entropy in the AC framework. They added a soft Q-network to the AC framework, where the Q-value from the soft Q-network is used to update the state-value network. The objective of this framework is not only to maximize expected reward but also to maximize entropy, encouraging the agent to explore a wider range of actions while giving up on obviously unpromising avenues.

### 2.2.4 Multi-agent DRL for MAPF

DRL algorithms can solve MAPF by treating it as a collection of multiple single-agent path planning problems, where each agent regards the others as part of the environment. While this approach simplifies the problem, it carries the risk of agents failing to address the non-stationarity and complexity of the environment, potentially resulting in collisions or deadlocks. Alternatively, a fully centralized approach can be employed, where DRL algorithms control the actions of all agents in a centralized manner, effectively mitigating issues related to environmental non-stationarity. However, this centralized approach becomes computationally expensive and less feasible for large-scale MAPF scenarios.

Multi-agent deep reinforcement learning (MADRL) offers an effective solution that combines the advantages of both approaches. In MAPF, the environment is not solely determined by a single agent's actions, as other agents also exert influence on the environment through their actions. MADRL leverages communication and coordination between agents to effectively address the complexity of the environments. Furthermore, MADRL enables agents to share knowledge with one another, following *centralized training and decentralized execution* paradigm (Gronauer and Diepold 2022). This approach streamlines the training process and accelerates learning by facilitating the exchange of mutual information (Foerster et al. 2018).

Similar to DRL algorithms, MADRL algorithms can be broadly divided into two groups: value-based methods and policy-gradient methods. A prominent example of a value-based method is Value Decomposition Networks (VDN) (Sunehag et al. 2017). VDN operates under the assumption that the team reward can be decomposed additively into individual Q-value functions for each agent. Its architecture learns based on the team reward, while individual actions are independently driven by each local observation. This approach encourages each agent to learn its own value function while simultaneously contributing to the overall team objective. Several extensions of VDN have been developed, further addressing the factorization of the global Q-value: QMIX (Rashid et al. 2020), QTRAN (Son et al. 2019, 2020) and QPLEX (Wang et al. 2020c).

QMIX builds upon the foundational framework of VDN but introduces a mixing network to incorporate individual Q-values in a non-linear fashion. The authors constructed the assumption that the global reward should adhere to a monotonic constraint between the global Q-value and individual Q-values. This empowers decentralized policies to make effective decisions based on both local observations and information received from neighboring agents. In contrast, QTRAN addresses this assumption of a monotonic constraint. They rather transform the global Q-value function into a factorizable one, demonstrating improved coordination among agents compared to VDN and QMIX. QPLEX also factorize the global value function by employing a duplex dueling network architecture. This network structure enhances the efficiency of value function learning and showcases superior performance over other value-based methods, particularly in micromanagement tasks.

On the other hand, Multi-Agent Deep Deterministic Policy Gradient (MADDPG) (Lowe et al. 2017) is one of the most noticed policy-gradient methods. It adapts the AC structure, employing centralized critics to train the actions and observations of all agents, while each agent has its decentralized actor that can access individual action-observations. MADDPG employs separate critic networks for each agent but shares network parameters. This separation allows agents to receive different rewards based on their observations and actions. Another approach, Counterfactual Multi-Agent (COMA) (Foerster et al. 2018), also utilizes the AC structure but introduces a counterfactual baseline that quantifies how much each agent's actions have contributed to the global reward compared to alternative actions they could have taken. COMA addresses the credit assignment problem by evaluating the contributions of each agent to the reward. A more recent policy-gradient method, Multi-Agent Proximal Policy Optimization (MAPPO) (Yu et al. 2022) incorporates a centralized value function input alongside Proximal Policy Optimization (PPO). In MAPPO, agents share both the policy network and critic network parameters. Additionally, it adopts Generalized Advantage Estimation (GAE) (Schulman et al. 2015b) as common practices for implementing PPO. Empirical findings by Yu et al. (2022) demonstrate that MAPPO achieves strong performance compared to other off-policy methods.

## 2.3 Evaluation for MAPF

Within this section, we aim to offer comprehensive evaluation indicators for MAPF algorithms. Clearly specifying the roles and definitions of indicators is crucial in assessing the performance of these algorithms, contributing to a more holistic evaluation framework.

We categorize evaluation indicators into two categories: quality indicators and indicators of the environmental complexity. Quality indicators measure the performance or effectiveness of an algorithm to assess how well an algorithm accomplishes its goal. On the other hand, Indicators of the environment's complexity provide the context regarding the difficulty and constraints of which the agents face in their environments to achieve their goal.

In the following subsections, we propose several unified indicators. In Sect. 2.3.1, we discuss quality indicators. In Sect. 2.3.2, we discuss indicators of the environment's complexity. Furthermore, In Sect. 2.3.3, we discuss limitations of indicator applicability.

### 2.3.1 Quality Indicators

The most widely recognized critical and mainly used quality indicator for evaluating the performance of different MAPF algorithms is *Success Rate*. It typically refers to the average percentage of agents successfully reaching the destination without any collisions, out of the total number of agents involved in given scenarios. However, some studies have included additional constraints into the *Success Rate* such as time limits (Sartoretti et al. 2019; Chan et al. 2022; Ma et al. 2021a, b; Wang et al. 2020; Chen et al. 2022a, b). Additionally, in some other literature, *Success Rate* is interpreted as the proportion of successful episode instances in which all agents reach their goals within a specified time limit (Semnani et al. 2020; Chen et al. 2022a, b). Although this quality indicator can provide valuable insights into both the efficacy and optimality of the MAPF algorithms, these diverse interpretations of the quality indicator cannot provide a unified evaluation of different algorithms.

First, we suggest adapting the *Success Rate* to represent the average percentage of agents successfully reaching the destination without any time constraints. In this definition, while it may consistently approach around 100% in less complex environments, it still provides information on how many agents made a collision or were unable to reach their goals by being blocked by other agents, particularly in complex environments. We also suggest evaluating the path length efficiency of the proposed algorithms with *Detour Percentage* (Wang et al. 2020). It provides a measure of the extent of how much deviation agents are making from their optimal path. It can be derived as follow:

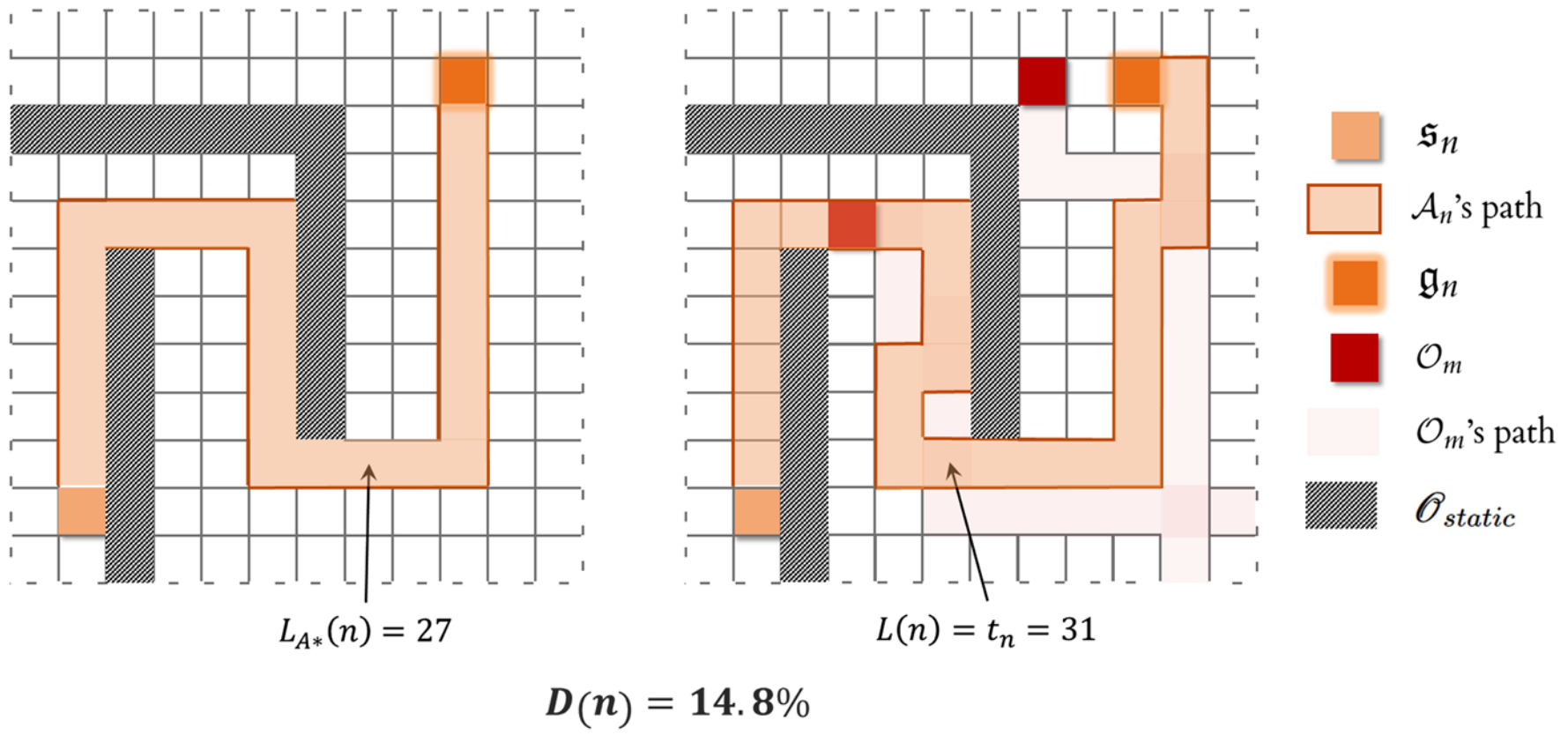

**Fig. 6** An example calculating *Detour Percentage D* after the agent $\mathcal{A}_n$ reaches its goal point in a grid map environment. $\mathfrak{s}_n$ and $\mathfrak{g}_n$ denote $\mathcal{A}_n$'s starting point and goal point, respectively. Dynamic obstacle $\mathcal{O}_m$'s location in the right figure shows the location at the time when $\mathcal{A}_n$ reaches its goal point. $\mathcal{O}_{static}$ denotes the set of static obstacles. The left-side figure shows the shortest path and the right-side figure shows the exact path $\mathcal{A}_n$ followed considering dynamic obstacles

$$\overline{D} = \frac{1}{N} \cdot \sum_{n=1}^{N} \frac{L(n) - L_{A*}(n)}{L_{A*}(n)} \times 100\%, \tag{16}$$

where $L(n)$ is the generated path's length of agent $n$, and $L_{A*}(n)$ is the agent $n$'s shortest path, which is calculated with the A* algorithm by only considering the single-agent path planning problem with static obstacles. In other words, it measures the additional distance that agents must travel due to the presence of other agents or dynamic obstacles. Figure 6 shows an example of calculating *Detour Percentage* in a given episode. The *Detour Percentage* is particularly useful in scenarios where agents must navigate through narrow passages or tight spaces, as it provides insight into the effectiveness of MAPF algorithms in such situations. Lower *Detour Percentage*s are generally desirable, as they indicate that the MAPF algorithm has not only minimized the impact of conflicts but also approached closer proximity to the optimal solution.

Lastly, we propose *Effective Velocity* as another quality indicator for evaluating time efficiency. It measures the effective distance that agents traveled within a single time step, disregarding path deviations caused by other agents or dynamic obstacles. It can be derived as follow:

$$\overline{v}_{eff} = \frac{1}{N^*} \cdot \sum_{n^*=1}^{N^*} \frac{L_{A*}(n^*)}{t_{n^*}}, \tag{17}$$

where $t_{n^*}$ represents the total time steps for agent $n^*$ to reach the goal. Figure 7 shows an example of calculating *Effective Velocity* in given episodes. Using $L(n^*)$ as the numerator term instead of $L_{A*}(n^*)$ would result in a constant $\overline{v}_{eff}$ (= 1 $grid/timestep$) in discrete environments, making it ineffective as a quality indicator. Using $L_{A*}(n^*)$ allows $\overline{v}_{eff}$ to measure the effective distance traveled within a single time step. In discrete environments, $\overline{v}_{eff}$ still aligns with $\overline{D}$ because agents maintain a constant speed for every time step, resulting in a proportional relationship between the total time step and the traveled distance. Hence, evaluating the time efficiency would be the same as evaluating the path length efficiency. However, in continuous environments, where agents' speed varies, $\overline{v}_{eff}$ effectively measures time efficiency.

### 2.3.2 Indicators of the environmental complexity

The evaluation of MAPF algorithms can vary depending on the constraints and conditions of the environment. Although the MAPF research community has made significant progress in creating benchmarks for MAPF, there is still room for improvement in properly defining the environmental complexity that reflects the difficulty of solving the MAPF problem instances. It is not sufficient to conclude that one algorithm performs better than another based solely on their performance in the same map. The results might be influenced by other factors in the environmental complexity. Therefore, it is essential to provide a clear measure of the problem's difficulty when evaluating algorithm performance. For example, recent work by Liu et al. (2020) introduced the MAPPER method, which outperformed PRIMAL (Sartoretti et al. 2019) in terms of *Success Rate*. However, it should be noted that MAPPER agents can move in eight directions, while PRIMAL* agents are constrained to only four. This difference in environmental flexibility may make it more difficult for PRIMAL agents to find paths, potentially impacting their performance. In this case, introducing an indicator of the environmental complexity that represents the utilization of space efficiency based on the degree of freedom in possible actions, would let us compare the difficulty of MAPF.

Therefore, we propose a new indicator of the environmental complexity, *Degree of Action Freedom* (*DAF*), which captures the different environmental constraints on agents' action options. This can be computed using the following equation:

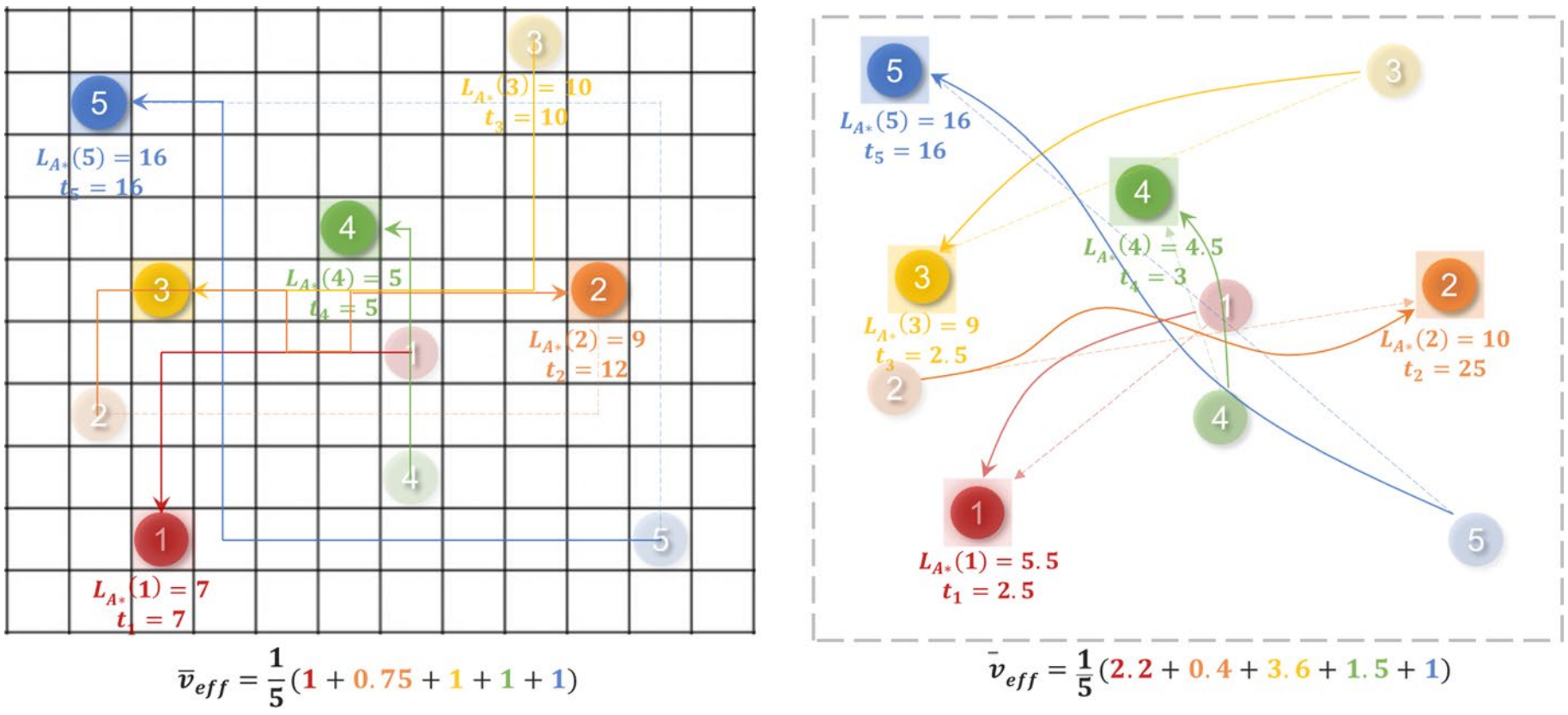

(a) Discrete Environment (b) Continuous Environment

**Fig. 7** An example of calculating *Effective Velocity*. The number of agents in both scenarios is five. In **a**, the agents can move a single grid in four directions for each time step. Whereas in **b**, the agents have the flexibility to move at varying speeds and in different directions

$$DAF = \frac{\mathsf{V}_{av}}{\mathsf{V}_{adj}}, \tag{18}$$

where $\mathsf{V}_{av}$ denotes all potential volumetric space that an agent can occupy after taking any action and $\mathsf{V}_{adj}$ denotes the volumetric space of all spatially adjacent space. Within this definition, $\mathsf{V}_{av}$ even includes the occupied space, which may result in a collision when the agent attempts to move. Figure 8 illustrates the concept of this indicator. For example, in PRIMAL (Sartoretti et al. 2019; Damani et al. 2021), each agent has five possible actions in a 2-dimensional map environment, where $DAF = 5/27 = 0.185$. In PRIMALc (Zhiyao and Sartoretti 2020), The agents have seven possible actions in a 3-dimensional map environment, where $DAF = 7/27 = 0.259$. This indicator reflects the range of action space of agents and provides insights into the flexibility and maneuverability of agents. Larger values of DAF indicate that agents can more efficiently deploy the space to choose their next actions, making it easier for them to avoid obstacles and reach their goals. By incorporating this indicator, we can better understand how difficult and restricted agents are in solving MAPF in different problem instances.

The performance of an algorithm can also vary depending on the conditions of problem instances. While an algorithm may successfully solve a MAPF problem with a large number of agents, its performance may drastically decline as the size of the map decreases and becomes more compact. This increased compactness will pose a greater challenge to solving the problem instance. Therefore, considering the level of congestion and crowding in the environment would provide insight into the difficulty of solving the problem. Therefore, we propose another indicator of the environmental complexity, *Crowdedness*, which quantifies the congestion in the environment. It is calculated using the following equation:

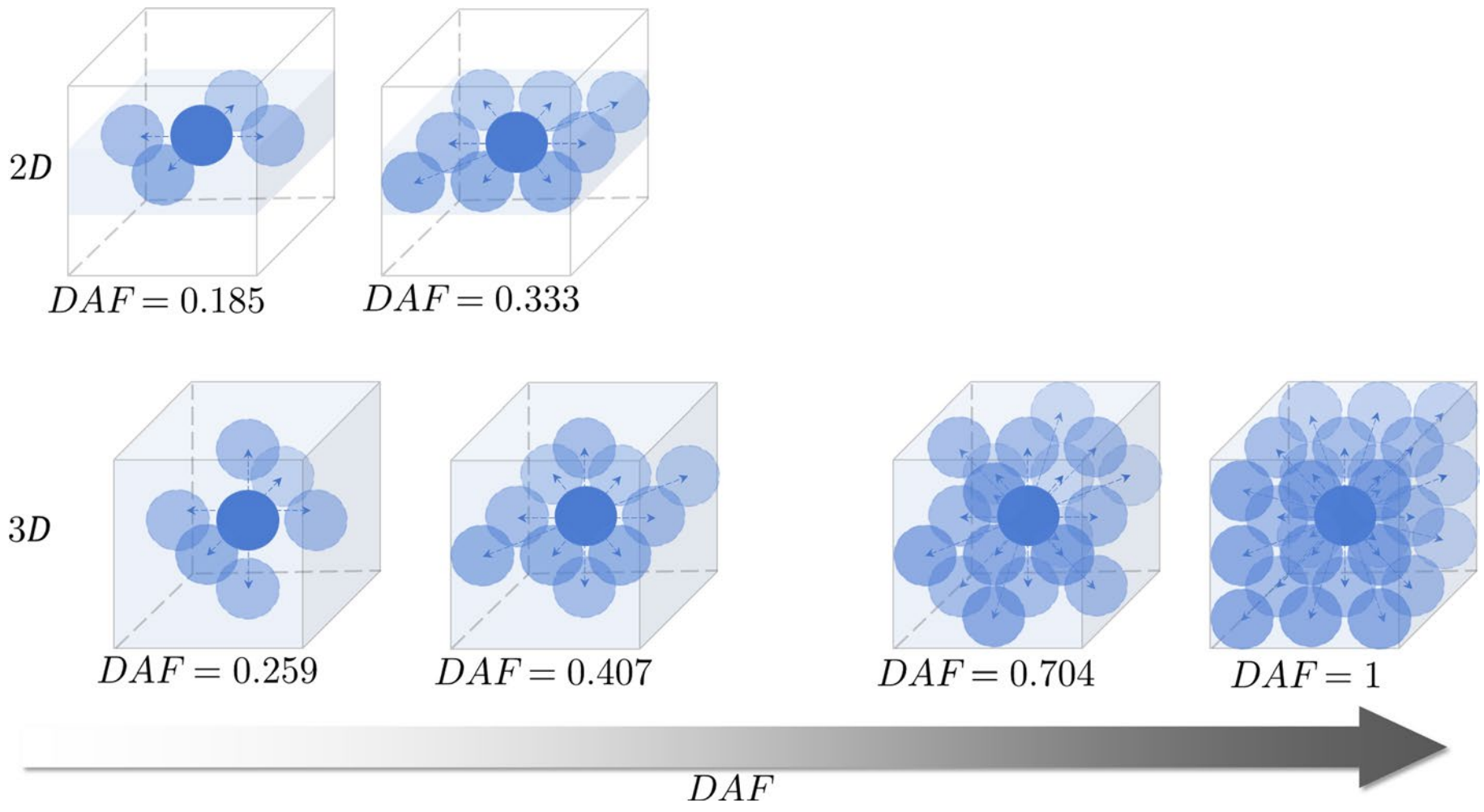

**Fig. 8** An illustration showing *DAF* in discrete space, depending on available choices for actions. The centered blue globe denotes the agent's current position and the blurred blue globe denotes the possible next position according to the choice of action options. Higher *DAF* indicates the agent has better maneuverability toward the environmental space allowing a lower difficulty in solving MAPF

$$Crowdedness = \frac{\Phi(\mathcal{A}) + \Phi(\mathcal{O}_{dyn})}{\Phi(\mathcal{V}) - \Phi(\mathscr{O}_{static})}, \quad (19)$$

where $\Phi(\cdot)$ is a function that derives the volume occupied by the variable input in the MAPF environment. In specific, it calculates the volume occupied by all agents $\Phi(\mathcal{A})$, the dynamic obstacles $\Phi(\mathcal{O}_{dyn})$, the entire map environment $\Phi(\mathcal{V})$, and the static obstacles $\Phi(\mathscr{O}_{static})$. For instance, in a grid map environment with $N$ agents and $M$ dynamic, and $K$ static obstacles, *Crowdedness* can be simply computed as $\frac{N+M}{|\mathcal{V}|-K}$. Figure 9 shows examples of calculating *Crowdedness* in 2*D* and 3*D* environments. A higher *crowdedness* indicates a more congested environment, which is expected to be more challenging for agents to navigate. In the work done by Guan et al. (2022), the performance results tend to exhibit improved performance as the map size increases and the number of agents and dynamic obstacles decreases. We believe a more comprehensive comparison of the performance between different algorithms would be available by evaluating different algorithms under the same level of *crowdedness*.

### 2.3.3 Limitations of indicator applicability

The indicators we proposed in previous Sects. 2.3.1 and 2.3.2 provide a generalizable evaluation framework for MAPF. Nonetheless, these indicators predominantly evaluate a single aspect of performance, potentially limiting the specialized assessment of an algorithm's capacity to handle specific scenarios. Recognizing these limitations of their applicability would help readers understand the scope of the evaluation. Quality indicators primarily focus on agents that successfully reach their goal points. However, these indicators lack an assessment of the agents that failed to reach the goals. For example, there is a challenge in determining how to count the agents blocked by other agents from reaching their destination when computing the Success Rate. The Detour Percentage and Effective Velocity indicators are also susceptible to missing a critical aspect of an algorithm's robustness and reliability in real-world scenarios. Furthermore, as each indicator is tailored to assess a specific facet of algorithm performance, it becomes difficult to blend both dimensions of assessment. For example, given all quality indicators discussed in Sect. 2.3.1, they may not perfectly present the holistic evaluation of algorithms performed in certain scenarios involving time limits, and distance constraints. These challenges emphasize the opportunity for further improvement in the evaluation framework, encouraging the development of more versatile metrics that can effectively assess algorithm performance across a wide range of scenarios

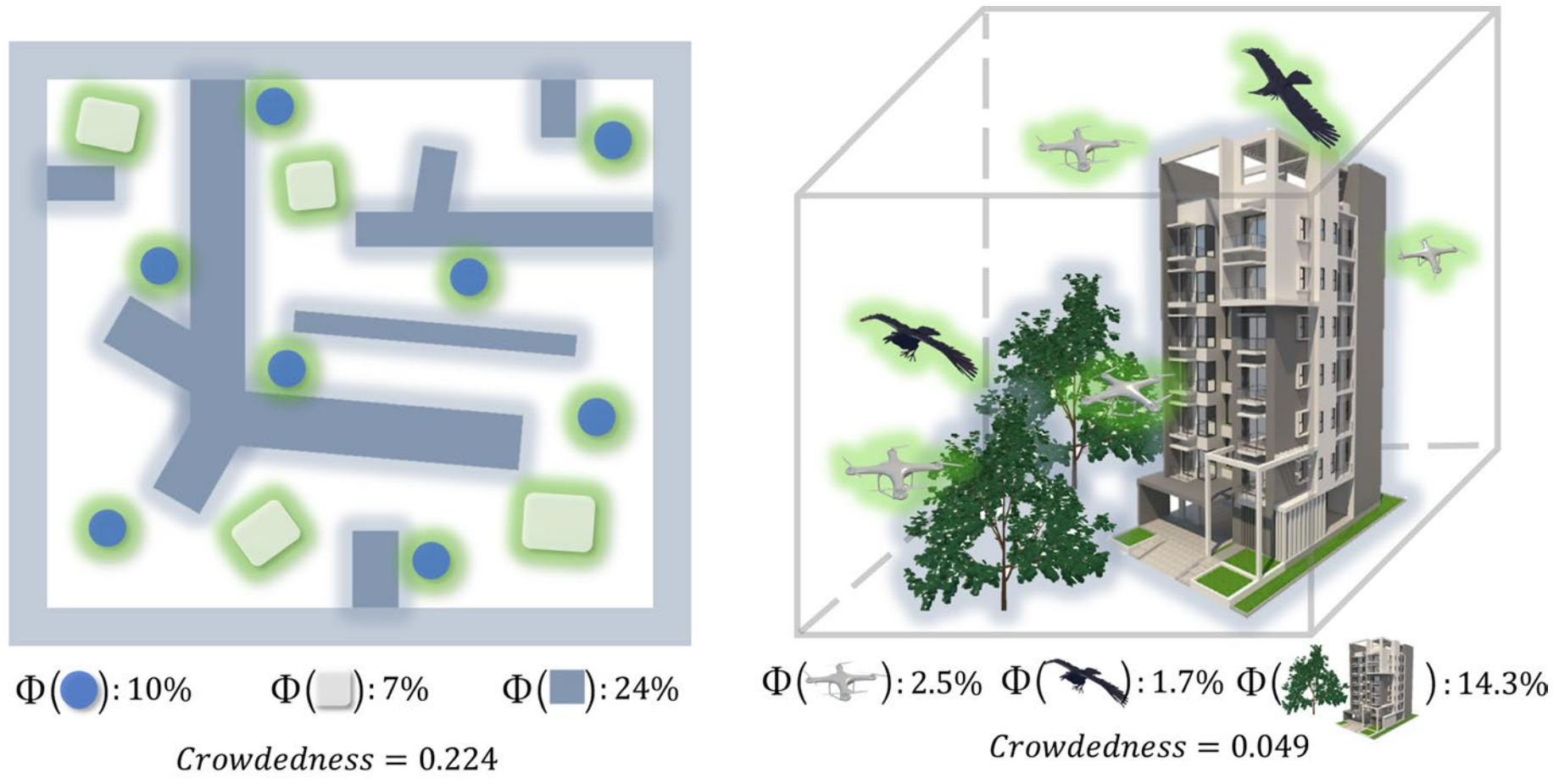

(a) Example of 2*D* Environment (b) Example of 3*D* Environment

**Fig. 9** Examples of calculating *Crowdedness* in 2D and 3D environments. In **a**, the blue circles and the gray rectangles denote the agents and dynamic obstacles, respectively, and the gray-blue walls denote the static obstacles. In **b**, the drones and the birds denote the agents and dynamic obstacles, respectively, and the trees and building denote the static obstacles

### 2.3.4 Simulation environments for DRL-based MAPF

Utilizing a unified environment for MAPF can facilitate comprehensive evaluations of MAPF algorithms as well. This is because when using the same MAPF environment, the rewards and computational costs for updating agents remain consistent, allowing for a fair comparison between different algorithms. Notably, there are two environment benchmarks that have recently gained prominence in RL community: VMAS (Bettini et al. 2022), and POGEMA (Skrynnik et al. 2022).

VMAS is a PyTorch-based vectorized 2D physics engine designed for simulating continuous multi-agent reinforcement learning environments. It offers diverse scenarios and is compatible with popular libraries such as OpenAI Gym, RLlib, and torchrl. Additional tutorials and installation details are available on the VMAS GitHub repository (VMAS github).

On the other hand, POGEMA is a grid-based MAPF simulator that targets partially observable environments with discrete spaces. It also provides an interface compatible with RL frameworks for both single-agent and multi-agent settings. We also refer the readers to found further information on the POGEMA GitHub repository (POGEMA github).

## 3 Recent DRL-based approaches for MAPF

In this section, we will discuss recent advancements in DRL-based approaches for MAPF within complex environments, which have demonstrated notable progress over the past five years. In recent years, there has been a significant increase in research dedicated to the development and advancements of model-free DRL approaches for MAPF. These novel methods have attempted to improve adaptability and performance in different environments where many issues are required to be considered due to complexity and uncertainty. As discussed in Sect. 2.2.3, model-free DRL algorithms can be categorized into two groups: value-based methods, and policy-gradient methods. In this study, our focus will primarily be on the recent value-based and policy-gradient model-free approaches. Table 1 shows the conditions and performance of model-free methods categorized according to their approaches.

### 3.1 Value-based approaches

DRL algorithms can provide improved solutions for MAPF by introducing prior knowledge to the network or improving the network structure. For example, Yang et al. (2020) proposed a method to address the slow convergence speed and excessive random action selection issue in the DQN algorithm. Prior to learning multiagent, they initialized the Q-network using prior knowledge obtained from a single automated guided vehicle in a static environment, utilizing the A* algorithm. This initialization process allowed the agents to acquire a better understanding of the environment, leading to improved learning efficiency. Additionally, they incorporated predefined prior rules to encourage the agents to avoid unnecessary exploration in the DQN algorithm. Additionally, Wang et al. (2020) demonstrated that a Duel neural network (Duel-NN) structure can outperform DQN and DDQN in solving the MAPF problem. Duel-NN differs from DQN and DDQN by incorporating an additional stream, the advantage function, which combines with the value stream to output the value function. This design helps the network effectively address the overestimation problem by reducing high gradient variance and bias. However, both approaches can still suffer from typical challenges commonly encountered by model-free methods, including long-range planning problems.

To address the long-range planning issue in MAPF, one potential solution is to design a novel reward structure that encourages desirable behaviors in specific scenarios that can cause collision or deadlock. Wang et al. (2020) proposed a globally guided RL approach with their novel reward structure that exploits spatio-temporal information of arbitrary environments and generalizes the solution. Their novel reward function was designed to integrate with the global path. This allows the agents to closely follow an efficient global path while effectively handling collision possible scenarios. Notably, this work was fully distributed in a non-communicative manner, demonstrating consistent performance across different environments regardless of the size and uncertainty. The success rate achieved by this work closely approached that of state-of-art centralized approaches. However, as environments grow increasingly complex, relying solely on manually and heuristically designed reward structures may not always guarantee the agents perfectly learn optimal decisions, particularly in unexpected or unfamiliar scenarios.

Another way to implicitly address the issue is by introducing communications between the agents. It can significantly reduce the non-stationarity caused by partial observations, alleviating the risk of collisions or deadlocks. Ma et al. (2021a) introduced Multi-Head Attention mechanism (Vaswani et al. 2017) and Gated Recurrent Unit (Chung et al. 2014) into communication block for the duel-DQN framework. This integration produced effective communication between agents and facilitated cooperation through graph convolution. Unlike previous works that relied on global guidance from single-agent shortest paths (Wang et al. 2020; Liu et al. 2020; Guan et al. 2022), their approach, known as 'Distributed, Heuristic and Communication' (DHC), introduced heuristic guidance via four-channel field of view, expanding the range of potential choices based on the embedding of shortest paths. DHC could successfully be applied to high-scale long-range environments. However, the use of broadcast communication in conjunction with graph convolution requires significant bandwidth and led to redundant information. Ma et al. (2021b) introduced the 'request-reply' communication scenario within the DHC framework to overcome this limitation, allowing agents to selectively communicate to focus only on relevant information for decision-making. Their extension of the framework, called 'Decision Causal

**Table 1** Overview of recent DRL-based approaches with respective conditions and performance results

| DRL Algorithm | | Paper | Decentralized Execution | Real-Robot Application | Existence of Dynamic Obstacles | Observability | Environment Space | Maximum Number of Agents Tested | Success Rate under Maximum Agents | Detour Percentage | Degree of Freedom | Crowdedness | Github link | Video link |
|---|---|---|---|---|---|---|---|---|---|---|---|---|---|---|
| Value-Based | DQN | Ma et al. (2021a) | ✓ | | | Part | Grid | 64 | 73 | – | 0.185 | 0.0143 | link | – |
| | | Ma et al. (2021b) | ✓ | | | Part | Grid | 128 | 82 | – | 0.185 | 0.0286 | link | – |
| | | Chen et al. (2022b) | ✓ | | | Part | Grid | 64 | 85 | – | 0.185 | 0.0571 | – | – |
| | DDQN | Wang et al. (2020) | ✓ | | ✓ | Part | Grid | 128 | 99.7 | 9.60 | 0.185 | 0.08 | - | link |
| | | Yang et al. (2020) | | | | Full | Grid | 8 | – | – | 0.185 | 0.0148 | – | – |
| | Dueling-DQN | Wang et al. (2020) | | | | Full | Cont | 4 | 92.3 | – | 0.333 | – | – | – |
| Actor-Critic | PPO | Long et al. (2018) | ✓ | | ✓ | Part | Cont | 20 | 96.5 | 9.50 | 0.333 | - | - | link |
| | | Fan et al. (2020) | ✓ | ✓ | ✓ | Part | Cont | 20 | 100 | 5.80 | 0.333 | 0.0354 | - | link |
| | | Wen et al. (2021) | ✓ | | ✓ | part | Cont | 4 | 96 | – | 0.333 | – | – | – |
| | | He et al. (2021) | ✓ | | ✓ | Full | Cont | 7 | 100 | – | 0.333 | – | – | – |

**Table 1** (continued)

| DRL Algorithm | Paper | Decentralized Execution | Real-Robot Application | Existence of Dynamic Obstacles | Observability | Environment Space | Maximum Number of Agents Tested | Success Rate under Maximum Agents | Detour Percentage | Degree of Freedom | Crowdedness | Github link | Video link |
|---|---|---|---|---|---|---|---|---|---|---|---|---|---|
| DDPG | Chen et al. (2020) | ✓ | | | Part | Cont | 4 | 92 | 0.333 | – | – | link | |
| | Qie et al. (2019) | | | | Full | Cont | 5 | – | – | 0.333 | – | – | – |
| | Guo et al. (2020) | ✓ | | | Part | Cont | 4 | – | – | 0.333 | – | – | – |
| | Hu et al. (2023) | | | | Full | Node | 5 | – | – | 0.185 | 0.037 | – | – |
| AC | Guan et al. (2022) | ✓ | | ✓ | Part | Grid | 150 | 99.1 | – | 0.333 | - | – | – |
| A2C | Liu et al. (2020) | ✓ | | ✓ | Part | Grid | 150 | 94 | – | 0.333 | – | – | – |

**Table 1** (continued)

| DRL Algorithm | Paper | Decentralized Execution | Real-Robot Application | Existence of Dynamic Obstacles | Observability | Environment Space | Maximum Number of Agents Tested | Success Rate under Maximum Agents | Detour Percentage | Degree of Freedom | Crowdedness | Github link | Video link |
|---|---|---|---|---|---|---|---|---|---|---|---|---|---|
| A3C | Sartoretti et al. (2019) | ✓ | ✓ | | Part | Grid | 1024 | 96 | - | 0.185 | 0.04 | link | link |
| | Damani et al. (2021) | ✓ | ✓ | | Part | Grid | 2048 | – | – | 0.185 | – | link | link |
| | Li et al. (2022b) | ✓ | | | Part | Grid | 64 | – | – | 0.185 | 0.08 | – | – |
| | Chan et al. (2022) | ✓ | | | Part | Grid | 1024 | 100 | – | 0.148 | 0.04 | link | – |
| | Semnani et al. (2020) | ✓ | | ✓ | Part | Cont | 16 | 100 | – | 0.333 | – | – | – |

The following abbreviations are used in the table: Part.: partially observable; Full.: fully observable map space; Cont.: continuous map space

Communication' (DCC), demonstrated improved results by effectively capturing relevant and influential neighbors for cooperation in MAPF.

Chen et al. (2022b) took a different approach by introducing a self-attention mechanism into their policy network to capture implicit collaborative information between agents from the observation. This makes their work distinctive from the works from (Ma et al. 2021a, b) because their agents make decisions reactively without explicit communication. To encourage agents to learn coordination, they developed hot supervision contrastive loss derived from an expert planner, which was combined with the RL loss function for the policy network. This combined loss function facilitated the agents' learning of effective coordination strategies.

### 3.2 Policy-gradient approaches

Taking advantage of initializing the learning process of DRL agents with prior knowledge about the environment dynamics, Guo et al. (2020) also demonstrated improved decision-making capabilities and faster convergence by incorporating the artificial potential field (APF) into DDPG algorithm. For mapless navigation, Hu et al. (2020) integrated DDPG with Prioritised Experience Replay (PER), leading to faster convergence, reduced fluctuation, and the generation of safer and smoother trajectories. However, as mentioned in the previous section, these approaches cannot fully address the issues coming from partial observability, particularly in high-scaled long-range planning problems.

Nonetheless, DDPG has been widely studied in multi-agent scenarios by centralized training and decentralized execution framework, so-called MADDPG. In MADDPG, a centralized critic learns non-stationary conditions, enabling decentralized actors for each agent to follow policies based on their private observation. In the automated container terminal scheduling for automated guided vehicles (AGVs), Hu et al. (2023) combined MADDPG with the Gumbel-Softmax strategy. Chen et al. (2020) applied MADDPG to their novel Delay-Aware Markov Game structure, alleviating performance degradation issues coming from delays as well as the non-stationary issue caused by the multi-agent system. For mapless navigation, Hu et al. (2020) integrated DDPG with Prioritised Experience Replay (PER), leading to faster convergence, reduced fluctuation, and the generation of safer and smoother trajectories. Another application of MADDPG was presented by Qie et al. (2019), who developed a simultaneous target assignment and path planning algorithm based on MADDPG. They combined target assignment rewards with path planning rewards to find the shortest distance while ensuring collision avoidance. This centralized training and decentralized execution framework can also be employed in other algorithms as well. For example, He et al. (2021) utilized asynchronous multithreading in PPO (AMPPO) for underwater unmanned vehicles application.

Another notably studied policy-gradient approach is PPO-based algorithms. Long et al. (2018) has extended MAPF to a multi-scenario end-to-end sensor-level decentralized MAPF solution in continuous space. Their approach involved mapping raw LiDAR sensor measurements to steering commands using a PPO, within the centralized learning and decentralized execution paradigm. To enhance the learning process, they incorporated the curriculum learning paradigm, inspired by Bengio et al. (2009), which accelerated the convergence to satisfactory solutions in a more efficient manner. While their work demonstrated successful multi-robot navigation in large-scale simulated multi-scenario environments, it could not provide satisfactory behavior when applied to real mobile robots. To bridge the simulation-to-real gap, they extended their work with a hybrid control framework that combines their prior RL model and traditional control schemes (Fan et al. 2020). This extension marked a significant advancement in validating the real-world applicability of DRL-based algorithms in MAPF, which had predominantly been explored in simulation-based applications. Similarly, Wen et al. (2021) introduced covariance matrix adaptation evolutionary strategies to the PPO algorithm as a means of enhancing the policy parameter update process. They further incorporated meta-reinforcement learning and transfer learning techniques within A3C framework. This approach enabled effective navigation of robots in unknown environments, achieving improved learning efficiency and reduced training time.

However, even with the advancements made by these DRL-based MAPF algorithms, challenges persist when dealing with long-range planning in large-scale congested environments. One of the possible approaches to address this is to decompose the problem into more manageable sub-problems. Liu et al. (2020) achieved this by using global waypoint coordinates as input to an A2C framework. By updating the networks based on evolutionary algorithm (Simon 2013), they were able to maintain good convergence properties, particularly in large-scale environments. The results demonstrated their method could greatly improve the DRL-based MAPF with a large number of agents, effectively handling dynamic obstacle avoidance. Building upon this work, Guan et al. (2022) further advanced the solution by incorporating a BicNet (Peng et al. 2017) into the design of the actor-network and introducing an attention mechanism to the critic-network. These enhancements contributed to improved stability and performance, resulting in the algorithm outperforming the MAPPER algorithm in comparative evaluations. These approaches signify promising directions for addressing the challenges of long-range planning in large-scale MAPF problems.

Another serious issue, deadlocks, pose a significant challenge for a number of agents, particularly in compact and highly structured environments such as maze environments. The rotation can be one main cause of deadlock as the agents' searching space increases from two dimensions to three dimensions. To address this specific deadlock scenario, Chan et al. (2022) developed a deadlock-breaking scheme that combines an imitation learning (IL) expert with A3C with a proper design of input maps and rewards. Through experimental results, the proposed approach demonstrated its efficiency in resolving MAPF problems with the existence of rotation movement. Another key challenge that can lead to deadlock scenarios is the application of a fully decentralized framework, as it makes it difficult to encourage each agent to act selflessly (Shoham and Leyton-Brown 2008). Li et al. (2022b) addressed this issue by adopting a hybrid approach that combines decentralized path planning with centralized collision avoidance. They developed a prioritized communication learning method (PICO), which integrates imitatively learned implicit planning priorities with a communication learning scheme. PICO has demonstrated its ability in success rates and collision rates, particularly in large-scale scenarios.

DRL-based MAPF algorithms also suffer from undesirable selfish behaviors of the agents. To address this issue, some researchers have proposed hybrid frameworks that combine RL with other techniques to handle critical scenarios where DRL-based policies may not make optimal decisions. For instance, Semnani et al. (2020) proposed a hybrid framework that switches policies between force-based motion planning and GPU/CPU A3C for Collision Avoidance with DRL (Everett et al. 2018), depending on the type of scenarios encountered by the robots. Skrynnik et al. (2023) proposed a hybrid policy that combines two policies, one derived from heuristic search and the other from RL, based on switch mechanism. Sartoretti et al. (2019) proposed a hybrid training framework that incorporates the 'Blocking Penalty' to prevent agents from pursuing their own goals while obstructing the paths of other agents. Additionally, they integrated an IL expert into the RL framework, combining off-policy behavior cloning with on-policy. To expose the agents to a diverse range of scenarios, the size and obstacle density of the map were randomized during the training process. The algorithm was successfully deployed on physical robots within a factory mockup, demonstrating their agents could learn collaborative behaviors. Damani et al. (2021) further upgraded PRIMAL to excel in highly structured and constrained worlds with lifelong tasks by learning behavior through conventions that improve implicit agent coordination.

Wang et al. (2023b)extended the prior work (PRIMAL) by incorporating a transformer-based communication learning mechanism into the PRIMAL framework. This integration effectively addresses the *chatter* problem, which arises due to conflicting messages from highly-scaled agents. This type of communication not only enables global information exchange among agents, facilitating cooperation, but also allows agents to gain knowledge about the historical observations of other agents, addressing the partial observability issue inherent in decentralized frameworks. Their study has demonstrated that this communication mechanism equips agents with sufficient information, even when their range of observation is highly limited. Additionally, He et al. (2023) employed a graph transformer architecture, enabling agents to access fuzzy global information and facilitating short-term predictions of agents' intentions. Their ablation study showed that this significantly enhances coordination between agents and substantially reduces episode length.

The aforementioned recent studies have made efforts to tackle the challenges of MAPF by integrating novel techniques that enable RL agents to gain a better understanding of their surroundings. These attempts provide the basic foundation for the transition to model-based DRL methods for MAPF, which will be discussed in Sect. 4.

## 4 Model-based DRL approach: a potential research frontier

We acknowledge that there are several gaps that need to be filled in MAPF. For example, addressing heterogeneous agents, Sim-to-Real applications, development of communication/coordination skills, credit-assignment problems, and robustness/safety concerns are all challenges that are currently under active research. These have been discussed in more details in other literature as future directions (Zhao et al. 2020; Zhang et al. 2021; Yakovlev et al. 2022; Gronauer and Diepold 2022; Wong et al. 2023). While there are different future directions in DRL-based MAPF approaches, in this section, we have chosen to focus our discussion on model-based DRL for MAPF, which is a promising direction that has not been sufficiently studied yet compared to other future directions. Despite the growing interest in the MAPF problem, it is notable that almost no attempts have been found that utilize model-based DRL methods to address the MAPF challenge. The relative novelty of the model-based approach, coupled with the inherent difficulties in acquiring reliable models of the environments, has limited the widespread adoption of such methods in MAPF.

Recently, Hafner et al. (2023) have presented a general and scalable model-based approach that successfully learns to achieve long-horizontal objectives in Minecraft without human intervention or heuristics. This work has sparked interest in exploring the use of model-based RL techniques in various domains (Taniguchi et al. 2023; Du et al. 2023; Wang et al. 2023a). Similarly, this model-based approach holds promise for addressing the complexity of large-scale and long-horizon MAPF scenarios, enhancing

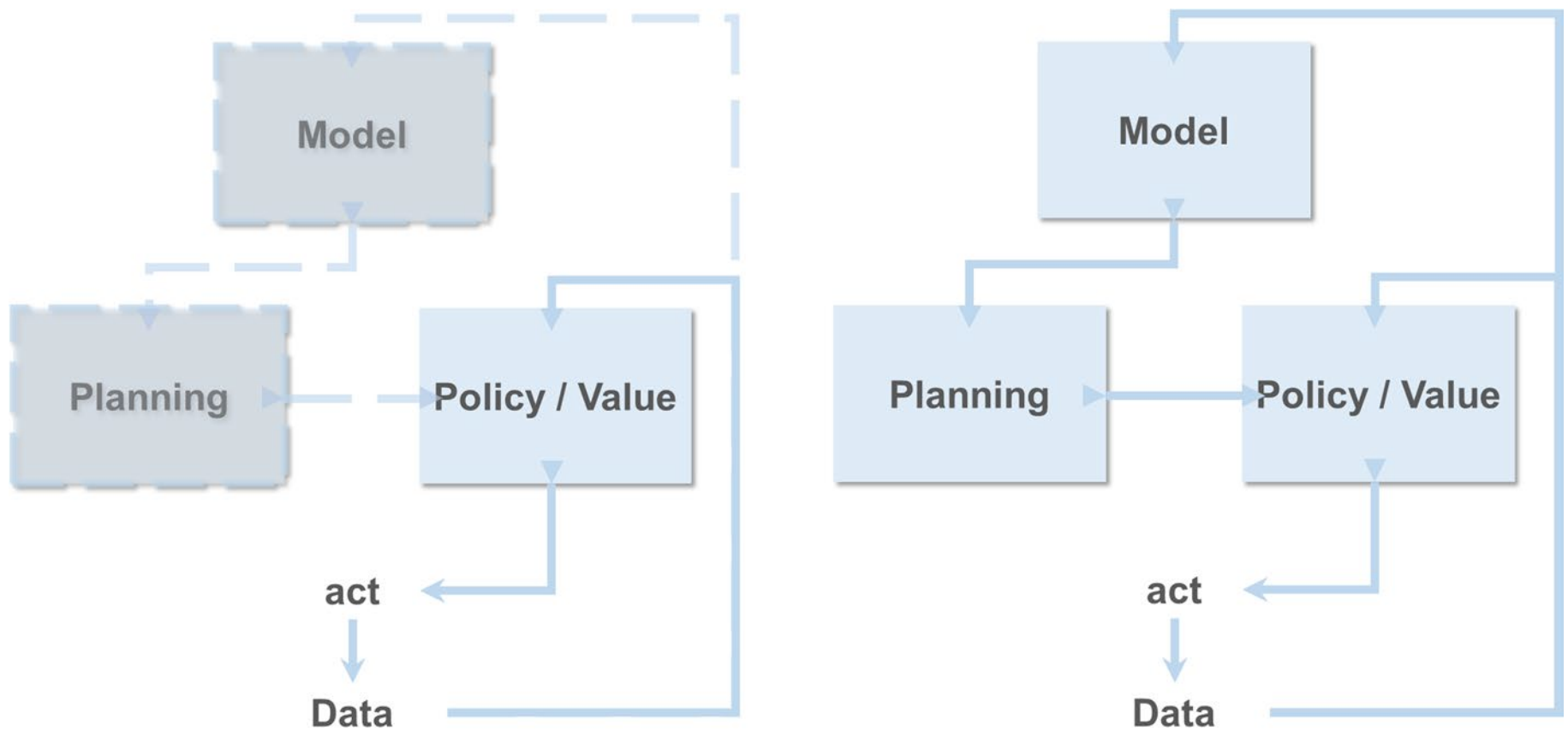

(a) Model-free method (b) Model-based method

**Fig. 10** Comparison of algorithmic flowchart between model-free method and model-based method. In **a**, the blurred boxes and blurred dashed arrows are not used

planning capabilities and improving learning processes. By combining DRL with explicit or learned environment models, model-based DRL enables agents to simulate and predict the consequences of their actions like proactive planners and hence facilitates informed decision-making and coordination.

Figure 10 represents the difference between the two frameworks of model-free methods and model-based methods. Model-free methods rely on past and current observations or states, employing one-step interaction rules, which often characterize them as reactive planners for MAPF (Wang et al. 2020; Everett et al. 2021; Chen et al. 2022b; Dergachev and Yakovlev 2021). On the other hand, model-based methods involve the preplanning stage with the model that has knowledge of the environment, allowing agents to make decisions within a broader context (Moerland et al. 2022). We, here, regard the model-based MAPF solver as a proactive planner, which integrates *i*) the future planning stage (Russell 2010) with a learned model and *ii*) a policy updating stage by using DRL to approximate global value or policy function in MAPF.

The model-based approach also has the potential to resolve the requirement of a tremendous number of samples for effective learning in MAPF. Some of its expected benefits are reduced exploration time with enhanced sample efficiency, improved understanding of complex dynamics, and adaptability to changing environments (Wang et al. 2022). As a result, it would likely advance and contribute to coordination and planning in MAPF. In the subsequent subsections, we will discuss two key steps for model-based approaches: the dynamics model learning stage and the integration of the planning and learning stage.

### 4.1 Dynamics model learning

The first key step of the model-based approach in MAPF is dynamics model learning. This step involves solving a supervised learning problem to determine the transition probabilities between states and associated rewards. In MAPF, given a batch of one-step transition as $\{s_t, a_t, r_t, s_{t+1}\}$ where $s_t$ is state, $a_t$ is action, $r_t$ is reward, and $s_{t+1}$ is the next state, the dynamics function can take on different forms (Moerland et al. 2023):

- Standard Markovian transition model: $(s_t, a_t) \rightarrow (s_{t+1}, r_t)$. Normally, the observation represents the state itself.
- Partial observability model: $((s_{t-l+1}, ..., s_t), a_t) \rightarrow (s_{t+1}, r_t)$. The lack of information in the current observation can be mitigated by considering the history of previous observations.
- Multi-step prediction model: $(s_t, (a_t, ..., a_{t+l-1})) \rightarrow (s_{t+l}, \sum_t^{t+l-1} r_t)$. Multi-step predictions by repeatedly applying one-step prediction models can accumulate errors because the one-step model is not learned to optimize long-term predictions. This may lead to inaccurate multi-step predictions from the true dynamics.
- State abstraction model: $(z_t, a_t) \rightarrow (z_{t+1}, r_t)$. This model learns the mapping between the observation and the compact latent representation, similar to auto-encoder (Rumelhart et al. 1985). State abstraction can reduce computational effort, and simplify planning and policy updating.
- Temporal/action abstration model: $(s_t, u_t) \rightarrow (s_{t+l}, \sum_t^{t+l-1} r_t)$. This idea identifies a high-level action space that spans multiple timesteps, directly implying the multi-step prediction model. This is commonly used in hierarchical RL, where the model learns an abstract action $u_t$.

These models are typically learned using neural networks that can handle high-dimensional inputs and effectively approximate non-linear functions across the entire state space.

Recently, there has been growing interest in the World Model presented by Ha and Schmidhuber (2018) due to its highly expressive nature and ability to learn rich spatial and temporal representations of data. In the World Model, the agent learns the dynamics of the environment by abstracting the high-dimensional data to a lower-dimensional one and only captures the significant features of the environment without any human intervention. In the context of MAPF, World Model can be used to abstract global information by modeling the entire context of agents based on partial observations. For instance, when modeling communication block, as presented in Wang et al. (2023b), or abstracting the prediction of agents' intentions, as presented in He et al. (2023), have a potential to enable higher-scaled scenarios. We believe applying this approach might be promising for enhancing the agents' ability to intelligently coordinate and understand the spatial context in MAPF.

## 4.2 Integration of the planning and learning

The next key step is the integration of the planning stage and policy updating stage. The planning stage involves determining the specifications for when and how to initiate planning, the duration of planning, and the actual planning process. For example, in the Dyna framework (Sutton 1991), planning is conducted only at previously visited states, and the model is sampled to generate 100 one-step transitions for each action derivation iteration. In the case of AlphaGo Zero (Silver et al. 2017), planning efforts are focused on the current state, with 1600 traces of Monte Carlo Tree Search (MCTS) (Coulom 2006) iterations conducted at a depth of 200.

The planning method also needs to be specified in this stage. The AI and RL communities have primarily utilized discrete planning method, such as probability-limited search (Lai 2015), breadth-limited depth-limited search (François-Lavet et al. 2019), and MCTS (Silver et al. 2017). However, there is also the option to employ differential planning methods, exploiting the differentiability of the learned model for a gradient-based approach. Examples of differential planning methods include iterative linear quadratic regulator planning (Bemporad et al. 2002), PILCO (Deisenroth and Rasmussen 2011), and DreamerV3 (Hafner et al. 2023, 2020, 2019). In this paper, we only provide a comprehensive framework for the planning methods, and for a more detailed review of recent planning methods, we refer readers to Moerland et al. (2020)'s paper.

Once the planning stage is configured, the next step is to establish the connection between the planning stage and the policy updating stage of learning. The output of the planning stage can be utilized to update the value or policy function of a model-free DRL network. In the predictive world model discussed in the previous subsection, the value or policy function behaves as a controller model, extracting useful representations of environmental information to guide the decision-making process with a small and simple neural network structure.

## 5 Conclusions

In this review, we provide a comprehensive overview of the current direction in MAPF, with a primary focus on deep reinforcement learning (DRL)-based approaches. We discuss the problem formulation of MAPF and recent approaches for solving it, and introduce the framework of DRL along with different templates of DRL algorithms applicable to MAPF. While numerous foundations and frameworks for solving MAPF have been established in the field, there is a relative lack of unified evaluation indicators for evaluating the performance of MAPF algorithms. In this study, we address this problem by proposing several metrics that can assess MAPF algorithms under diverse conditions and facilitate performance comparisons between different solutions. Furthermore, classical DRL algorithms suffer from long-range planning problems and they may not guarantee generalized solutions in highly non-stationary environments. We investigate how recent DRL techniques have tackled these challenges in high-scale and complex MAPF environments. Lastly, we establish a basic understanding of model-based DRL approaches and conclude that they can offer a breakthrough to further advance MAPF solutions.

**Author contributions** JC: Conceptualization, Methodology, Formal Analysis, Investigation, Data Curation, Visualization, Writing—Original Draft. JF: Conceptualization, Writing—Review & Editing. YA-Y: Conceptualization, Writing—Review & Editing. HN: Supervision, Writing—Review & Editing

## Declarations

**Conflict of interest** The authors declare no conflict of interests.